\documentclass[a4paper,fleqn]{cas-dc}

\usepackage[authoryear,longnamesfirst]{natbib}
\usepackage{times}
\usepackage{epsfig}
\usepackage{graphicx}
\usepackage{amsmath}
\usepackage{amssymb}

\usepackage{multirow}
\usepackage{color}
\usepackage{hyperref}

\def\tsc#1{\csdef{#1}{\textsc{\lowercase{#1}}\xspace}}

\tsc{WGM}
\tsc{QE}

\begin{document}

\let\WriteBookmarks\relax
\def\floatpagepagefraction{1}
\def\textpagefraction{.001}

\shorttitle{CCNet for Multi-spectral Vehicle Re-identification and A High-quality Benchmark}    

\shortauthors{\textit{Aihua Zheng ~et al.}}  

\title [mode = title]{Cross-directional Consistency Network with Adaptive Layer Normalization for Multi-spectral Vehicle Re-identification and A High-quality Benchmark}  



%

\author[1]{Aihua~Zheng}[
orcid = 
]



\ead{ahzheng214@foxmail.com}


\credit{Conceptualization of this study and Methodology}

\affiliation[1]{organization={Information Materials and Intelligent Sensing Laboratory of Anhui Province, Anhui Provincial Key Laboratory of Multimodal Cognitive Computation, School of Artificial Intelligence},
            addressline={Anhui University }, 
            city={Hefei},
            postcode={230601}, 
            country={China}}

\author[2]{Xianpeng~Zhu}[
orcid =
]


\ead{xpzhu6325@foxmail.com}


\credit{ Investigation and Writing-Original Draft}

\affiliation[2]{organization={Anhui Provincial Key Laboratory of Multimodal Cognitive Computation, School of Computer Science and Technology, },
            addressline={Anhui University}, 
            city={Hefei},
            postcode={230601}, 
            country={China}}

\author[2]{Zhiqi~Ma}[
orcid =
]


\ead{doermzq0398@foxmail.com}


\credit{Validation and Visualization}

\author[1]{Chenglong~Li}[
orcid = 0000-0002-7233-2739
]


\ead{lcl1314@foxmail.com}


\credit{Formal Analysis and Data Curation  }

\author[2]{Jin~Tang}[
orcid =
]


\ead{tangjin@ahu.edu.cn}


\credit{Resources and interpretation of data}

\author[3]{Jixin~Ma}[
orcid =
]


\ead{j.ma@greenwich.ac.uk}


\credit{Writing Review and Editing}

\affiliation[3]{organization={School of Computing and Mathematical Sciences},
            addressline={University of Greenwich, London SE10 9LS, UK}, 
            city={London},
            postcode={}, 
            country={UK}}

\cortext[1]{Corresponding Author: Chenglong Li. Xianpeng Zhu and Zhiqi Ma are with equal contributions.}



\begin{abstract}
To tackle the challenge of vehicle re-identification (Re-ID) in complex lighting environments and diverse scenes, multi-spectral sources like visible and infrared information are taken into consideration due to their excellent complementary advantages.
		However, multi-spectral vehicle Re-ID suffers cross-modality discrepancy caused by heterogeneous properties of different modalities as well as a big challenge of the diverse appearance with different views in each identity.
		Meanwhile, diverse environmental interference leads to heavy sample distributional discrepancy in each modality.
		In this work, we propose a novel cross-directional consistency network \textcolor{red}{(CCNet)} to simultaneously overcome the discrepancies 
		from both modality and sample aspects.
		In particular, we design a new cross-directional center loss \textcolor{red}{($L_{cdc}$)} to pull the modality centers of each identity close to mitigate cross-modality discrepancy, while the sample centers of each identity close to alleviate the sample discrepancy. Such a strategy can generate discriminative multi-spectral feature representations for vehicle Re-ID.
		In addition, we design an adaptive layer normalization unit \textcolor{red}{(ALNU)} to dynamically adjust individual feature distribution to handle distributional discrepancy of intra-modality features for robust learning.
		To provide a comprehensive evaluation platform, we create a high-quality RGB-NIR-TIR multi-spectral vehicle Re-ID benchmark (MSVR310), including 310 different vehicles from a broad range of viewpoints, time spans and environmental complexities.
		Comprehensive experiments on both created and public datasets demonstrate the effectiveness of the proposed approach comparing to the state-of-the-art methods.
		The dataset and code will be released for free academic usage at \textcolor{blue}{\url{https://github.com/superlollipop123/Cross-directional-Center-Network-and-MSVR310}}.
\end{abstract}


\begin{highlights}
\item A cross-directional consistency network for multi-spectral vehicle re-identification. 
		
\item A cross-directional center loss to simultaneously pull modality and sample centers.
				
\item An adaptive layer normalization to adjust feature distribution in each modality. 

\item A high-quality multi-spectral vehicle re-identification benchmark dataset MSVR310.

\end{highlights}

\begin{keywords}
 Vehicle Re-ID\sep Multi-spectral Representation\sep Cross-directional Center Consistency\sep Layer Normalization\sep Benchmark Dataset
\end{keywords}

\maketitle

\section{Introduction}\label{}
        \begin{figure}[t]
		\centering
		\includegraphics[width=0.99\columnwidth]{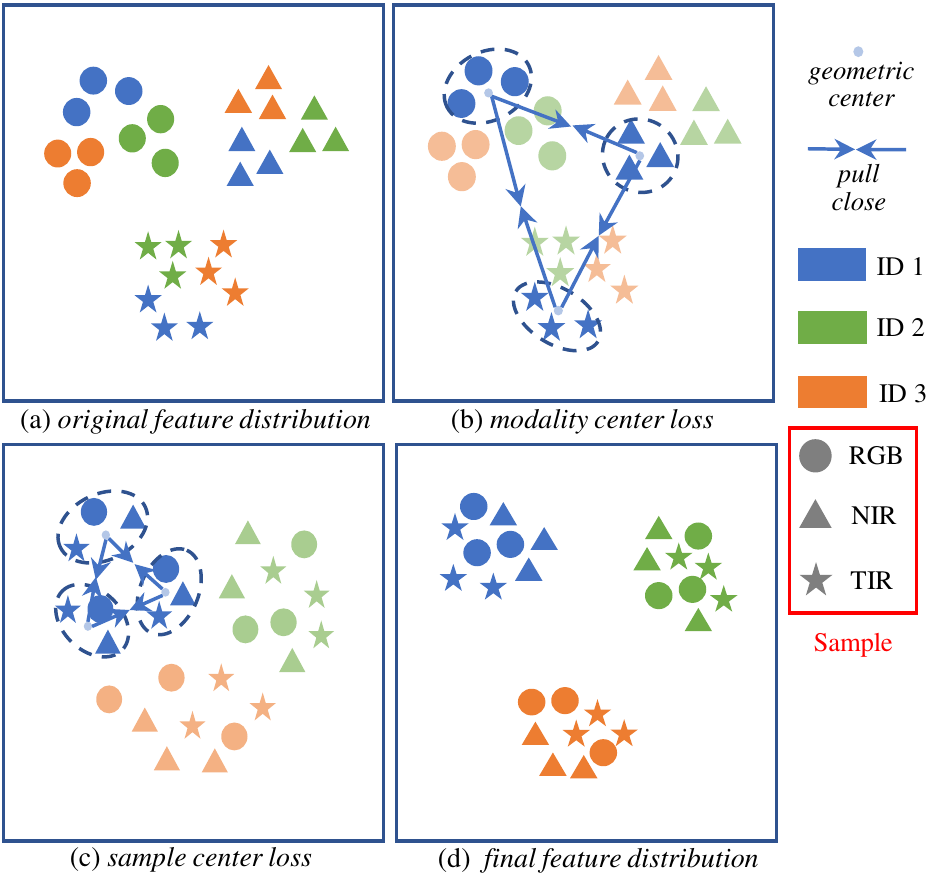}
		\caption{
			Impact demonstration of the Cross-directional Center loss $\mathcal{L}_{CdC}$ on the distribution of multi-spectral features.
			(a) The original distribution.
			(b) Impact of $\mathcal{L}_{CdC_M}$ which aims to pull modality centers of each identity close.
			(c) Impact of $\mathcal{L}_{CdC_S}$ which pulls sample centers of each identity close.
			(d) Feature distribution driven by $\mathcal{L}_{CdC}$ including both $\mathcal{L}_{CdC_M}$ and $\mathcal{L}_{CdC_S}$. 
		}
		\label{fig:cdc_loss}
        \end{figure}

Vehicle re-identification (Re-ID) aims to search the given vehicle image from the cross-camera gallery with the same identity.
	Due to the wide range of real-life applications such as video surveillance, smart city and intelligent transportation.
	Vehicle Re-ID has been attracted growing attention and experiencing rapid development with the emergence of comprehensive studies~\cite{Chu2019VehicleRW, Lou2019EmbeddingAL, Tang2019CityFlowAC, AN2022136} and public large-scale datasets~\cite{Liu2016LargescaleVR, Liu_2016_CVPR, Lou2019VERIWildAL,guo2018learning}.
	However, most existing studies only focus on visible images which suffer imaging weaknesses in complex lighting environments and extreme weather, thus can not satisfy the demand for all-day and all-weather real-life surveillance.

	Since visible (RGB), near infrared (NIR) and thermal infrared (TIR) sources have strongly complementary advantages in adverse lighting conditions and environments.
	RGB-NIR-TIR multi-spectral vision tasks, such as tracking~\cite{Lu2021RGBTTV,Li2016LearningCS,Tu2022M5LMM, AFYOUNI2022279}, person Re-ID~\cite{zheng2021robust} and saliency detection~\cite{Tu2021MultiInteractiveDF} attract hot research interest in the both machine learning and computer vision communities. 
	Recently, Li \textit{et al.}~\cite{li2020multi} first launch the multi-spectral vehicle Re-ID task.
	They first propose a baseline multi-spectral vehicle Re-ID method Heterogeneity-Collaboration Aware Multi-Stream Convolutional Network (HAMNet) which utilizes multi-spectral features with class-aware weight fusion.
	Meanwhile, they first provide two benchmark datasets RGBN300 and RGBNT100 to multi-spectral vehicle Re-ID community. 
	%
	%
	To be annotated, different from traditional vehicle Re-ID datasets which treat a single image as a sample, these two multi-spectral datasets treat an image pair (RGB-NIR in RGBN300) or an image triplet (RGB-NIR-TIR) as a sample.
	\textcolor{blue}{To avoid confusion, we use the concept \textit{sample} in the rest of this paper to emphasize the difference from conventional single modality \textit{image} for multi-modal Re-ID.}
	Despite of the pioneer contribution, there are three major issues remain to be well addressed in multi-spectral vehicle Re-ID.
	
	First, the sample discrepancy caused by the diverse imaging conditions and the modality discrepancy with the heterogeneous modality gap restrict the learning capacity of intra-class compactness.
	We propose a cross-directional center loss $\mathcal{L}_{CdC}$ which is composed of \textcolor{red}{a sample center loss} $\mathcal{L}_{CdC_S}$ and \textcolor{red}{a modality center loss} $\mathcal{L}_{CdC_M}$ to solve the sample and modality discrepancies from multi-modality aspect.
	On the one hand,
	it is hard to distinguish identities by only a certain spectrum data under complex environmental interference while the modality gap significantly disturbs directly utilization of different modality.
	To reduce the heterogeneous gap while taking the advantages of consistent information among modalities, we propose to enforce the centers of images with the same ID from different modalities in a mini-batch close by introducing a modality center loss $\mathcal{L}_{CdC_M}$, as shown in Fig.~\ref{fig:cdc_loss} (b). 
	In this way, we can intuitively enforce the modality consistency and reduce the disturbance caused by a certain modality image.

	On the other hand, 
	although sample relation has been widely concerned in RGB and cross-modality retrieval task by triplet loss~\cite{Chu2019VehicleRW,Hermans2017InDO}, center loss~\cite{2016A}, HC loss~\cite{2020Hetero},
	cross-modality constrains~\cite{Ye2018VisibleTP,ye2019bi} and metric learning~\cite{ling2020class}, they are not suitable for complementary and heterogeneous multi-modality images. 
	%
	The heavy environmental interference caused by illumination challenge ubiquitously exists in multi-modality data. In this case, a certain image from a certain modality is possibly unreliable when it suffers from extreme environmental interference, and will easily introduce abnormal relation in pair-wise metric process.
	\textcolor{blue}{For instance, Ling \textit{et al.}~\cite{ling2020class} strengthens the relational constraints between the modality center and the samples by metric learning on the intra-class, inter-class, and intra-modal and inter-modal relationships for cross-modality Re-ID. 
	However, it relies on features from a single image when learning intra-modal relations, which is sensitive to the noise within a certain modality, similar to center loss~\cite{2016A}, \textcolor{red}{MAUM~\cite{maum}} and \textcolor{red}{HRNet~\cite{wu2022end}}. 
	Meanwhile, it is optimized by constraining the distances between positive and negative sample pairs, resulting in the optimization of the intra-class relationship relying on the inter-class relationship.
	Therefore it cannot well contain the intra-class difference since the distribution of positive and negative samples in multi-modality vehicle re-identification is extremely complex. }

\textcolor{blue}{By contrast, our cross-directional center (CdC) loss takes the intra-class consistency relationship as the goal to optimize the stronger multi-modal intra-class relationships. At the same time, CdC loss does not rely too much on single image features during optimization, which can reduce the interference of low-quality images in a certain modality during the training process.
HC loss~\cite{2020Hetero} solves the problem of modal differences in cross-modality Re-ID by constraining the distance between the centers of different modality features within the class. 
While our CdC loss constrains both the intra-class modality discrepancy and the sample discrepancy, which can better overcome the huge intra-class variance in multi-modality vehicle data. }

	Therefore, to learn more robust features from the complementary multi-modality images, we propose a sample center loss to pull the centers of each triplet (RGB-NIR-TIR) sample with the same identity in a mini-batch close in this paper, as shown in Fig.~\ref{fig:cdc_loss} (c).
	%
	%
	By jointly optimizing sample center loss and modality center loss in a cross-directional fashion (as shown in Fig.~\ref{fig:network}) in a unified deep learning framework, it simultaneously reduces both intra-class sample discrepancy and  cross-modality heterogeneity, as shown in Fig.~\ref{fig:cdc_loss}.
	%

	Second, multi-modality data are usually collected in diverse and challenging environments where single modal data can not satisfy the demand for robust recognition.
	In this case, the data style and quality is complex which increases the difficulty of learning relations from every single modality.
	Meanwhile, diverse environmental interference and large appearance gap also disturb the  process of identity consistency relation learning.
	Therefore, due to the diverse environmental interference, features from single modality suffer from heavy distributional variation, as shown in Fig.~\ref{fig:ALNU_cmp}. This increases the difficulty in robust feature learning for CNN and further impacts the intra-class identity consistency learning.
 
\textcolor{blue}{	To reduce the disturbance of intra-modality distributional variation, we design a simple but effective module called adaptive layer normalization unit (ALNU), {to normalize the individual features and adaptively adjust their distributions without breaking their inner relations.}}
	%
{Different from existing normalization operations like BN (\textcolor{red}{Batch Normalization})~\cite{2015Batch}, 
 IN (\textcolor{red}{Instance Normalization})~\cite{ulyanov2016instance} and GN (\textcolor{red}{Group Normalization})~\cite{Wu2018GroupN}, ALNU treats each input feature as an entirety and preserves original information without changing the relation across channels in feature when adjusting the distribution.
	Comparing with traditional layer normalization (LN)~\cite{ba2016layer} which also doesn't change the relations across channels, our ALNU adaptively learns the gain and bias factors according to original inputs by introducing extra convolution and pooling layers and thus is more flexible.}
	%
	%
	Specifically, we integrate ALNU into all branches in our network to greatly improve the discriminative ability of multi-spectral target representations and thus further boost the performance of multi-spectral vehicle Re-ID.
	
	
	\begin{figure}[t]
		\centering
		\includegraphics[width=0.99\columnwidth]{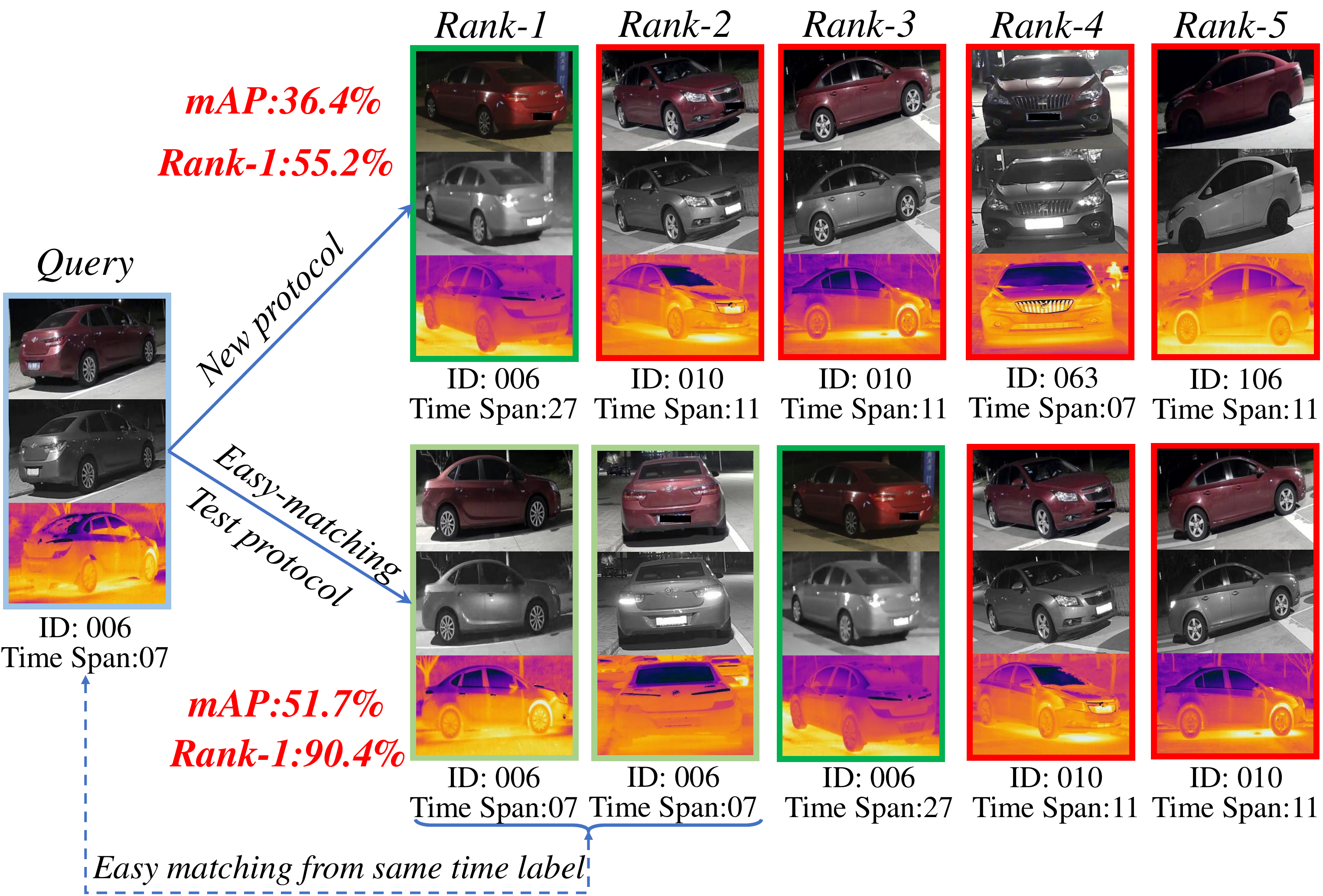}
		\caption{
			Illustration of the comparison of the results between the proposed test protocol and the easy-matching protocol. {\textcolor{red}{Since the easy matching may easily hit the samples from the same camera or with the same viewpoint across different time spans, it results in 15.3\% and 35.2\% higher in mAP and Rank-1 respectively.}}
            }
		
		\label{fig:test_protocol}
	\end{figure}
	
	Third, existing multi-spectral vehicle Re-ID datasets, RGBN300~\cite{li2020multi} and RGBNT100~\cite{li2020multi}, are limited in diversity. 
	To provide a more comprehensive evaluation platform in multi-spectral vehicle Re-ID, we create a high-quality image benchmark dataset named MSVR310.
	Compared with the RGBNT100 dataset~\cite{li2020multi}, our MSVR310 has following two benefits.
	
	
	{\bf \flushleft Longer time span}. MSVR310 is collected across a relative long time span (over 40 days). Benefiting by the long time span, data collected in MSVR310 have various environmental conditions such as various illuminations, occlusions and weather. It effectively increases the diversity of our dataset. Furthermore, we annotate the time labels of samples according to their collection sequences along time. These labels would be used in improving the experimental evaluation of multi-spectral vehicle Re-ID.
	
	{\bf \flushleft More reasonable protocol}. Although most advanced methods forbid to match the samples from the same camera such as Market1501~\cite{Zheng2015ScalablePR}, VeRi-776~\cite{Liu2016LargescaleVR}, or the same viewpoint such as in RGBNT100, RGBN300~\cite{li2020multi} to avoid the easy matching, it is not realistic enough since the same vehicle may appear in the same camera or with the same viewpoint across different time spans. 
	Therefore, we propose to prevent the easy matching caused by similar identity-unrelated information like environments and noises by a more reasonable label, and time span, instead of the camera/viewpoint as the new protocol. Fig.~\ref{fig:test_protocol} shows the easy matching protocol in RGBNT100 with the same time span, even though with the different viewpoints, the vehicles with the same identity and time label can be easily distinguished from others due to their high similarity on image content.

	As summary, we propose a end-to-end Cross-directional Consistency Network (CCNet) to simultaneously overcome modality and sample discrepancies. And propose a new multi-spectrum vehicle Re-ID dataset MSVR310 with diverse illustration interference and rich view variation with more reasonable protocol.
	The contributions of this paper can be summarized as follows.
	\begin{itemize}
		\item We propose a novel cross-directional consistency network based on the cross-directional center loss to simultaneously address the problems of cross-modality discrepancy caused by heterogeneous properties of different modalities and intra-class appearance discrepancy caused by different views and adverse lighting conditions in multi-spectral vehicle Re-ID. 
		
		\item We propose an adaptive layer normalization unit to dynamically {adjust feature distribution within each modality}. We integrate the unit into each modality branch in our network to help reducing the disturbance of intra-modality distributional variation.

		\item We create a high-quality benchmark dataset MSVR310, including 310 different vehicles from a broad range of viewpoints, time spans and environmental complexities. The benchmark will provide a comprehensive evaluation platform to promote the research and development of multi-spectral vehicle Re-ID. 
		
		\item Comprehensive experiments on our dataset MSVR310 and the public dataset RGBNT100 validate the superior performance of our approach against several state-of-the-art multi-spectral vehicle Re-ID methods.
		We also conduct a random modality-missing experiment to prove the robustness of CCNet in facing the issue of missing modalities.
	\end{itemize}

 \section{Related Work}
	
	We briefly review the related works in vehicle Re-ID, cross-modality person Re-ID and multi-modality person Re-ID.
	
	\subsection{Vehicle Re-ID}
	
	In last few years, vehicle Re-ID has gained a growing attention with the rapid development of Re-ID task, which boosts the development of intelligent cities~\cite{WEN2014130}.
        %
	Liu \textit{et al.}~\cite{Liu2016LargescaleVR} propose a dataset called VeRi-776 with a coarse-to-fine progressive searching framework using multiple information like license plate and spatio-temporal label.
	Liu \textit{et al.}~\cite{Liu_2016_CVPR} release another large-scale vehicle Re-ID dataset VehicleID and build a distance related method.
	Some works~\cite{Wang2017OrientationIF,Shen2017LearningDN} introduce spatio-temporal information to provide a stricter constraint besides  utilization of normal visual features. 
	%
	VANet~\cite{Chu2019VehicleRW} propose a metric loss function by treating vehicle image pairs with same or not same viewpoints differently to acquire a better distance measure.
	He \textit{et al.}~\cite{He2019PartRegularizedNV} design a method to enhance discriminative feature representation by introducing detection methods.
        \textcolor{red}{Li~\cite{li2022attribute} propose to embed attributes and state information to enhance feature learning by reducing the intra-class feature gap.}
	Khorramshahi \textit{et al.}~\cite{Khorramshahi2019ADM} introduce key-points information to utilize adaptive attention for vehicle Re-ID.
	Semantic segmentation~\cite{Meng2020ParsingBasedVE} is utilized to split feature into different parts with corresponding regions in vehicles, followed by a part-aligned metric way to measure distance of image pairs more precisely.
	Recently, more large-scale and challenging vehicle datasets are released, like VERI-Wild~\cite{Lou2019VERIWildAL} and CityFlow~\cite{Tang2019CityFlowAC}.
	Besides real data, synthetic dataset~\cite{Yao2020SimulatingCC} constructed via graphic engine emerges to provide arbitrary environments for learning.
	However, all these methods mentioned above only take a usage of single RGB modality, which is hard to satisfy the demand for all-day all weather monitoring over long period.
	
	
	\subsection{Cross-Modality Person Re-ID}
	To handle illumination limitations in RGB-based person Re-ID, Wu \textit{et al.}~\cite{Wu2017RGBInfraredCP} propose the first RGB-Infrared cross-modality benchmark SYSU-MM01 and a deep zero-padding network.
	RegDB~\cite{Nguyen2017PersonRS} is also a widely used cross-modality dataset with paired visible and thermal images collected by dual camera system.
	Ye \textit{et al.}~\cite{Ye2018VisibleTP} suggest a two-stream network with triplet loss to constrain the similarity in cross-modality images.
	An effective loss~\cite{2020Hetero} is designed to supervise network learning modality invariant feature by constraining the intra-class center distance in modalities.
	Ye \textit{et al.}~\cite{ye2019bi} propose a bi-directional center-constrained loss to handle cross-modality and intra-modality variations simultaneously.
	Wang \textit{et al.}~\cite{Wang2019RGBInfraredCP} introduce a generating model to translate images to opposite modality to acquire pixel level alignment and make a feature level constraint with joint discriminator to push network produce discriminative features.
	Li \textit{et al.}~\cite{2020Infrared} introduce an auxiliary intermediate modality to reduce the gap between modalities.
	Lu \textit{et al.}~\cite{Lu2020CrossModalityPR} propose a novel cross-modality shared-specific feature transfer algorithm to explore both modality-shared and modality-specific information.
        Huang \textit{et al.}~\cite{crxmodinfofus} provided a comprehensive and detailed review for cross-modality person re-id and outline the future research trends.
        \textcolor{red}{Wei \textit{et al.}~\cite{Wei_2021_ICCV} propose to incorporate features of heterogeneous images to generate modality-invariant representations.}
        \textcolor{red}{Ye \textit{et al.}~\cite{ye2021dynamic} propose a dynamic tri-level relation mining framework to explore intra-modality and cross-modality relations.}
        \textcolor{red}{Wei \textit{et al.}~\cite{wei2021flexible} propose a flexible body partition model-based adversarial learning to enhance feature discriminability.}
        \textcolor{red}{Wei \textit{et al.}~\cite{wei2022rbdf} propose a reciprocal bidirectional framework for modality unification and discriminative feature learning.}
	However, due to the lack of real aligned paired images in modalities, the heterogeneous issue in cross-modality person Re-ID still remains a key challenge.
	%
	
	\subsection{Multi-Modality Person Re-ID}
	Similar to infrared images, depth images do not suffer the influence on lighting variation and can reflect shape and distance information of targets.
	Barbosa \textit{et al.}~\cite{Barbosa2012ReidentificationWR} first propose RGB-D person Re-ID with a corresponding dataset named PAVIS.
	Mgelmose \textit{et al.}~\cite{Mgelmose2013TrimodalPR} combined three different information including RGB, depth and thermal data in a joint classifier, which is the first time to utilize RGB, depth and thermal sources in person Re-ID.
	Munaro \textit{et al.}~\cite{Munaro20143DRO} collect a RGB-D dataset named BIWI with 50 identities and multiple data sources.
	Wu \textit{et al.}~\cite{Wu2017RobustDP} utilize depth data to provide more invariant body shape and skeleton information  to overcome change of illumination and color.
	A new cross-modality distillation network~\cite{Hafner2018ACD} has been proposed to transfer supervision between modalities like similar structural features and make a discriminative mapping to a common feature space.
	However, depth information is difficult to be utilized in outdoor open environments which seriously limits its application in this task.
	
	To provide a robust solution for overcoming environmental interference, Li \textit{et al.}~\cite{li2020multi} first launch multi-spectral vehicle Re-ID datasets RGBN300 (visible and near infrared) and RGBNT100 (visible, near infrared and thermal infrared), and propose a baseline method {HAMNet~\cite{li2020multi} to effectively learn better feature representation by class-aware weight fusing and consistency prediction constraining.
	However, HAMNet~\cite{li2020multi} mainly focuses on learning multi-modality feature relations and ignores the discrepancy in both sample and modality levels. Our CCNet mainly focuses on mitigating the widely existed discrepancies from both modality and sample aspects by introducing cross-directional center loss.}
	Zheng \textit{et al.}~\cite{zheng2021robust} release a new multi-spectral person Re-ID dataset RGBNT201, and a progressive fusion network for multi-modality fusion.
        Chen \textit{et al.}~\cite{CHEN2023445} designed a model to inherit the advantages of CNN and Transformer for multimodal matching.
	Although these two works first launch RGB-NI-TI multi-spectral Re-ID task and provide two benchmark datasets and baseline methods for vehicle and person Re-ID respectively, how to effectively fuse the complementary but heterogeneous information is still a big challenge.

\section{Cross-directional Consistency Network}
		\begin{figure*}
		\centering
		\includegraphics[width=0.99\textwidth]{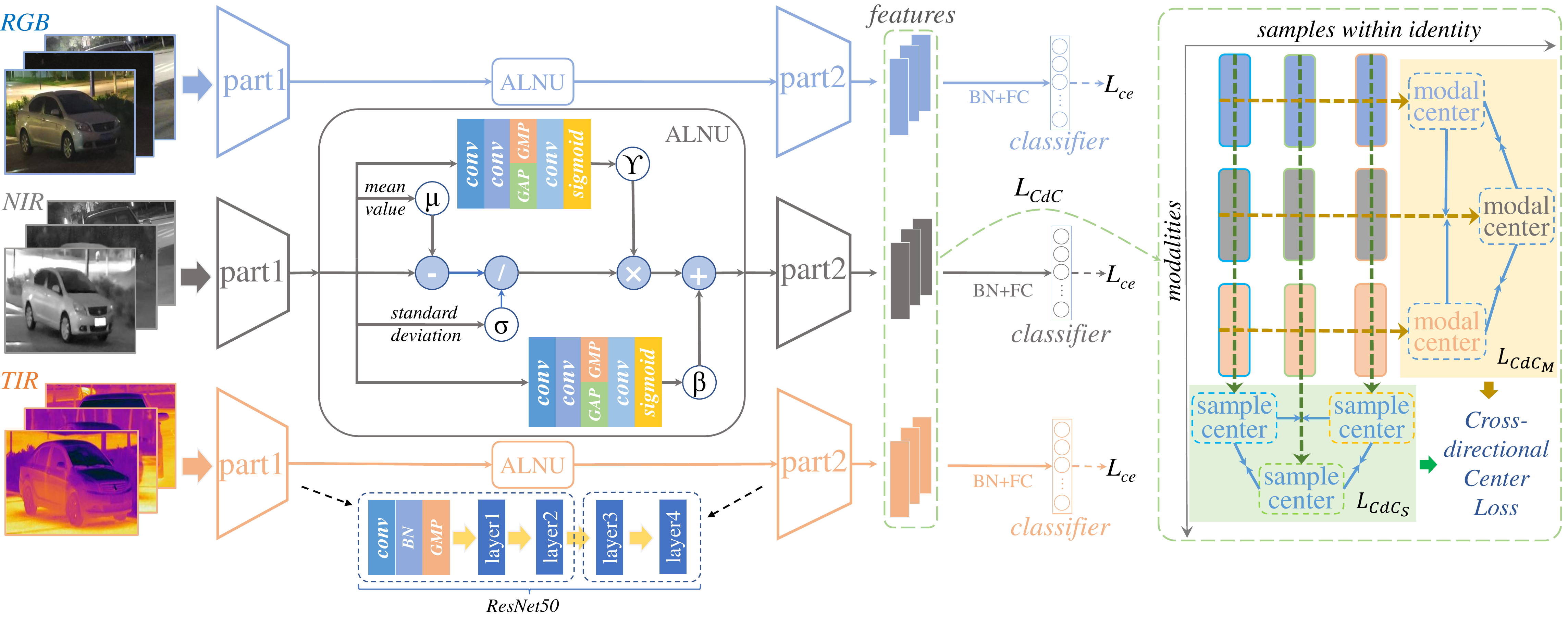}
		\caption{
            Framework of the proposed Cross-directional  Consistency Network (CCNet). A multi-stream network is designed to handle RGB, NIR and TIR data separately at the body part with an adaptive layer normalization unit (ALNU) embedded at the middle for each branch. Each branch is an independent ResNet50 which is split into two parts at the position between its layer2 and layer3. Then CdC loss is utilized to mine the potential intra-class relation in sample and modality level.
            }
        \label{fig:network}
	\end{figure*}
	
	To utilize the consistency and mitigate the discrepancy in multi-spectral data, we propose a robust method with cross-directional center loss and adaptive layer normalization unit for multi-spectral vehicle Re-ID, referred as Cross-directional Consistency Network (CCNet) in this paper.
	
	As shown in Fig.~\ref{fig:network}, CCNet is a multi-branch structure with three equivalent branches aiming to extract specific features for each single spectral data. 
	Given a sample with multiple modalities, we send the image from each spectrum into corresponding branch without sharing the parameters.
	In each branch, an individual ALNU (adaptive layer normalization unit) module is integrated at the middle to modify feature distribution.
	For input images in training mini-batches, their features are divided into different groups according to the identity.
	Then cross-directional center loss is introduced to mitigate the intra-class appearance discrepancy and cross-modality discrepancy simultaneously for multi-spectral vehicle Re-ID.
	Each branch makes a prediction supervised by the cross entropy loss to learn the identity related features.

	In this work, we use 
 $D = \{ I_i \mid 1 \le i \le N \}$
 donating the whole dataset where $N$ is the identity size.
	$I_i = \{ S_{i,n} \mid 1 \le n \le N_i \}$
 donates the sample set belonging to the $i^{th}$ vehicle where $N_i$ is the sample number of the vehicle $I_i$.
	$S_{i,n} = \{ x^m_{i,n} \mid 1 \le m \le M \}$
 donates the image set of 
 $n^{th}$
 sample from 
 $I_i$
 and 
 $x^m_{i,n}$
 is the single image from the 
 $m^{th}$
 modality in the sample 
 $S_{i,n}$
 .
	In this work, $M$ is $3$ and we can simply donate samples in a triplet form as $S_{i,n} = (x^1_{i,n},  x^2_{i,n}, x^3_{i,n})$ to represent images from RGB, NIR and TIR modality respectively.
	%
        We use $Part^m_k$ to donate the $k^{th}$ part of the branch for the $m^{th}$ modality in CCNet.
	Then, the forward process for the image $x^m_{i,n}$ can be formulated as:
	\begin{equation}
		\begin{aligned}
                {f^m_{i,n} = Part^m_2({ALNU}^m(Part^m_1(x^m_{i,n}))),}
		\end{aligned}
	\end{equation}
	where $f^m_{i,n}$ donates the correspond feature for the image $x^m_{i,n}$. And the final representation for $S_{i,n}$ is the concatenation of its corresponding feature triplet $(f^1_{i,n}, f^2_{i,n}, f^3_{i,n})$.

	\subsection{Adaptive Layer Normalization Unit}

\begin{figure}[t]
		\centering
		\includegraphics[width=0.99\columnwidth]{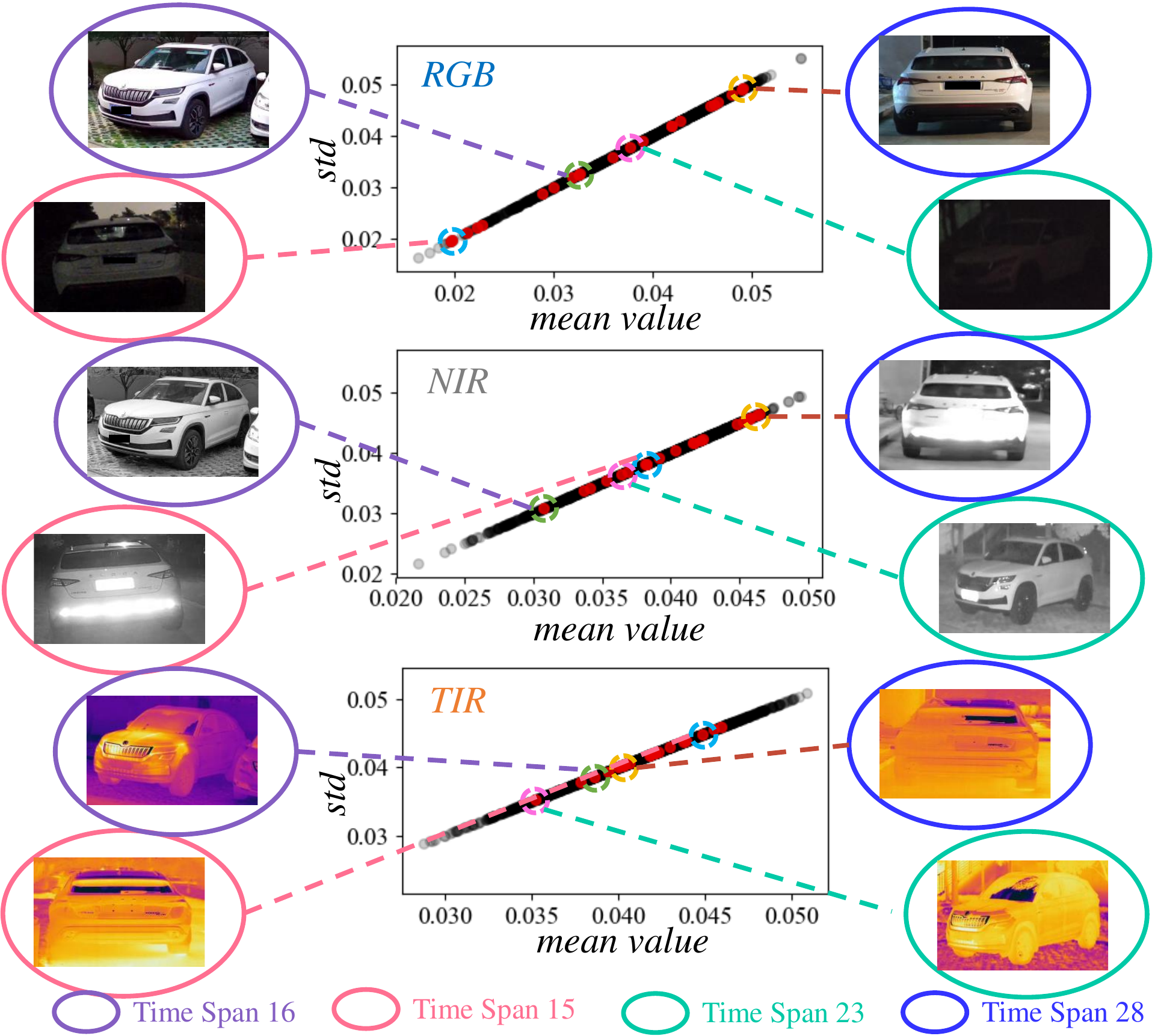}
		\caption{
  	         \textcolor{red}{The heavy feature distributional variation issue of the multi-modality images with the same ID in four different time spans.}
			The red points together with the example images are with the same identity, which scatters with large discordance. The three images with the same color oval boxes indicate the RGB, NIR and TIR images collected in the same time span. The black points correspond to the features of the images in MSVR310 dataset.
		}
		\label{fig:ALNU_cmp}
	\end{figure}

	{ALNU aims to handle heavy feature distributional variation within a certain modality caused by sample differences and complex environmental interference. Specifically, it normalizes the individual features and adaptive adjusts their distributions without breaking their inner relations.}
	\textcolor{blue}{As shown in Fig.~\ref{fig:ALNU_cmp}, the mean value and standard deviation of intra-modality features are distributed in a wide range, even the images with the same identity from the same modality, which further influence the intra-class identity consistency learning.
	ALNU module tries to mitigate the disturbance caused by heavily distributional variation by normalizing each input feature and adjusting the distribution dynamically.}
	On one hand, this operation reduces the discrepancy on distribution of intra-modality features and helps to extract more robust CNN features.
	On the other hand, it is hard to evaluate similarity accurately for intra-modality images with large distribution gap regardless of identity. And mitigating this discrepancy helps to improve the validity of final similarity comparing of intra-modality image pairs in multi-spectral vehicle Re-ID task.
		
	%
	Given an input image $x^m_{i,n}$, we acquire its middle feature before sending into ALNU as:
	\begin{equation}
		\begin{aligned}
			{f^{i,m,n}_{mid}}= Branch^m_1(x^m_{i,n}),
		\end{aligned}
	\end{equation} 
	where ${f^{i,m,n}_{mid}}$ is a 3-D tensor with the shape of $H$, $W$, $C$. 
	We can easily obtain its mean value and standard deviation as:
	\begin{equation}
		\begin{aligned}
			\mu = \frac{1}{HWC}\sum_{h=1}^{H}\sum_{w=1}^{W}\sum_{c=1}^{C} {f^{i,m,n}_{mid}},
		\end{aligned}
	\end{equation}
	\begin{equation}
		\begin{aligned}
			\sigma = \sqrt{\frac{1}{HWC}\sum_{h=1}^{H}\sum_{w=1}^{W}\sum_{c=1}^{C}({f^{i,m,n}_{mid}} - \mu)}.
		\end{aligned}
	\end{equation}

	Then, we calculate a normalized feature:
	\begin{equation}
		\begin{aligned}
			\hat{f}_{mid} = \frac{f_{mid} - \mu}{\sqrt{\sigma^2 + \epsilon}},
		\end{aligned}
	\end{equation}
	where $\epsilon$ is a small value to avoid the division over zero.
	%
        \textcolor{red}{BN~\cite{2015Batch} first propose to rescale and shift features in Batch Normalization, however the scale factors and shift factors are fixed during inference stage. By contrast, we propose to adaptively parameterize these two factors according to input data during both training and inference stage.} 
	Each ALNU module contains two adaptive learning blocks (${ALB}_{\gamma}$ and ${ALB}_{\beta}$), each of which is stacked by two convolutional layers, two parallel pooling layers, another convolution layer and a $Sigmoid$ activation function. ALNU dynamically acquires two extra scalars by two adaptive learning blocks according to original input $f_{mid}$ to further adjust the distribution of normalized feature $\hat{f}_{mid}$.
	This process can be formulated as:
 
	\begin{equation}
		\begin{aligned}
			f^{\prime}_{mid} = \hat{f}_{mid} \odot \gamma + \beta,
		\end{aligned}
	\end{equation}
 
	where $\gamma = ALB_{\gamma}(f_{mid})$, $\beta = ALB_{\beta}(f_{mid})$, and $f^{\prime}_{mid}$ is the final output of ALNU.
	
  Compared to conventional normalization operations like BN~\cite{2015Batch}, IN~\cite{ulyanov2016instance}, which adjust the original feature distribution in channel level, ALNU module works for individual features without breaking the relation among inner channels to avoid distinct change of the original feature distribution.
	Compared with LN~\cite{ba2016layer} which enforces features to follow the same mean value and variance in evaluation, our ALNU learns the gain and bias factors $\gamma$ and $\beta$ from original input features to adaptively adjust the distribution.
	Different from conventional normalization operation like BN~\cite{2015Batch}, LN~\cite{ba2016layer}, GN~\cite{Wu2018GroupN} which help models to learn easier and faster, ALNU mainly focus on intra-modality distributional variation for features, which is unrelated to their identity and increases the difficulty in robust feature learning.
	On one hand, ALNU adaptively modifies the distribution of features within modality and reduce the discrepancy caused by environmental interference which further mitigates the disturbance of identity related information learning.
	On the other hand, ALNU adaptively learns the gain and bias factors for each feature to achieve more flexible adjustment instead of enforcing all features to follow identical mean value and variance.

	\subsection{Cross-directional Center Loss}

	Compared with single spectral data, multi-spectral ones include more information but more challenges in vehicle Re-ID data.
	The challenges can be mainly summarized from two aspects, including sample discrepancy and modality discrepancy.
	For the sample discrepancy, a suitable representation for sample to satisfy the form of multi-modality data is necessary.
	Meanwhile, ubiquitous bad cases from a certain modality in multi-modality data have to be taken into consideration.
	For the modality discrepancy, the heterogeneous gap among modalities prevents the direct utilization for multi-modality data.       
	We propose cross-directional center loss $L_{CdC}$ to handle above discrepancies and mine a better identity embedding in multi-spectral vehicle Re-ID.
         \textcolor{red}{The proposed $L_{CdC}$ not only considers the relation between modalities like HC loss \cite{2020Hetero}, but also take the relation between samples into consideration, as shown in Fig.~\ref{fig:cdc_loss}.}
	%
	

	In training process, we randomly select $P$ identities with $K$ samples in each mini-batch, which forms totally $M\times K \times P$ images.
	Then, let $F_i = \{ f^m_{i,k} \mid 1 \le m \le M, 1 \le k \le K \}$ donate the final features belonging to the $i^{th}$ identity in a training mini-batch.
	The geometric sample center for the $k^{th}$ sample in $F_i$ can be formulated as:
	\begin{equation}
		\begin{aligned}
			{C_S}_{i,k} = \frac{1}{M} \sum_{m=1}^{M} f^m_{i,k}.
		\end{aligned}
	\end{equation}
	
	To overcome the sample discrepancy in multi-modality case, we propose a \textcolor{red}{Sample Center Loss} to pull intra-class sample centers as close as possible. 
	This process can be formulated as:
	\begin{equation}
		\begin{aligned}
			\mathcal{L}_{CdC_S} =
			\frac{1}{2K(K-1)}
			\sum_{i=1}^P
			\sum_{1 \le k_1 < k_2 \le K} \left\|{C_S}_{i,k_1} - {C_S}_{i,k_2} \right\|^2_2.
		\end{aligned}
	\end{equation}
	
	Similar, the geometric modality center for the $m^{th}$ modality in $F_i$ can be formulated as:
	\begin{equation}
		\begin{aligned}
			{C^m_M}_{i} = \frac{1}{K} \sum_{k=1}^{K} f^m_{i,k}.
		\end{aligned}
	\end{equation}
	
	In the same manner, to overcome the modality discrepancy, we propose a \textcolor{red}{Modality Center Loss} to pull intra-class modality centers as close as possible. This process can be formulated as:
	\begin{equation}
		\begin{aligned}
			\mathcal{L}_{CdC_M} =
			\frac{1}{2M(M-1)}
			\sum_{i=i}^P
			\sum_{1 \le m_1 < m_2 \le M} \left\|{C^{m_1}_M}_{i} - {C^{m_2}_M}_{i} \right\|^2_2.
		\end{aligned}
	\end{equation}
	
	Then, the cross-directional center loss $L_{CdC}$ is defined as:
	\begin{equation}
		\begin{aligned}
			\mathcal{L}_{CdC} = \mathcal{L}_{CdC_S} + \mathcal{L}_{CdC_M}.
		\end{aligned}
	\end{equation}
	
	More intuitive demonstration is shown in Fig.~\ref{fig:network}.
	The gradients of $L_{CdC}$ with respect to $f^m_{i,k}$ can be solved as (since $L_{CdC}$ only concerns intra-class relation, we simply ignore $i$ below):
	\begin{equation}
		\begin{aligned}
			&\frac{\partial L_{CdC}}{\partial f^m_k}
			=
			\frac{\partial L_{CdC_S}}{\partial f^m_k} + \frac{\partial L_{CdC_M}}{\partial f^m_k} \\
			=
			&\frac{1}{K-1}({C_S}_k - \bar{C_S}) \frac{\partial {C_S}_k}{\partial f^m_k} 
			+
			\frac{1}{M-1}({C^m_M} - \bar{C_M}) \frac{\partial {C^m_M}}{\partial f^m_k} \\
			=
			&\frac{1}{M(K-1)}({C_S}_k - \bar{f}) + \frac{1}{K(M-1)}({C^m_M} - \bar{f}),
		\end{aligned}
		\label{equ:gradient}
	\end{equation}
	where $\bar{C_S}$, $\bar{C_M}$, $\bar{f}$ can be formulated as:
	\begin{equation}
		\begin{aligned}
			\bar{C_S} = \frac{1}{K} \sum_{k=1}^{K} {C_S}_k
			= \frac{1}{MK} \sum_{k=1}^{K} \sum_{m=1}^{M} f^m_k
			= \bar{f},
		\end{aligned}
	\end{equation}
	\begin{equation}
		\begin{aligned}
			\bar{C_M} = \frac{1}{M} \sum_{m=1}^{M} {C^m_M}
			= \frac{1}{MK} \sum_{m=1}^{M} \sum_{k=1}^{K} f^m_k
			= \bar{f}.
		\end{aligned}
	\end{equation}
	
	Thus, the final optimizing strength of $L_{CdC}$ with respect to $f^m_{k}$ is linearly dependent on its corresponding sample center ${C_S}_k$, modality center $C^m_M$ and global identity center $\bar{f}$.
	Intra-class features within sample (modality) are in same gradient along sample (modality) direction.
	Besides, the gradient of $L_{CdC}$ with respect to $f^m_k$ is not directly related with $f^m_k$ itself, which is not such sensitive when $f^m_k$ corresponds to the bad cases in a certain modality.

	In this work, $K$ and $P$ is set to $4$ and $8$ respectively.
	As shown in Eq.~\eqref{equ:gradient}, the final factors of gradient along sample and modality directions are different ($\frac{1}{M(K-1)}$ and $\frac{1}{K(M-1)}$ respectively).
	Thus, we introduce a hyper-parameter $\alpha$ to balance their strengths. The final formulation of $L_{CdC}$ is defined as:
	\begin{equation}
		\begin{aligned}
			\mathcal{L}_{CdC} = \mathcal{L}_{CdC_S} + \alpha \mathcal{L}_{CdC_M}.
		\end{aligned}
		\label{eq:L_cdc}
	\end{equation}

	Cross-directional center loss $L_{CdC}$ focuses on optimizing intra-class relation along sample and modality directions. To enhance the ability of discriminative inter-class learning, we further introduce the cross entropy loss $L_{ce}$.
	The total loss is defined as:
	\begin{equation}
		\begin{aligned}
			\mathcal{L}_{total} 
			=  
			L_{ce}
			+
			\lambda L_{CdC},
		\end{aligned}
		\label{eq:L_total}
	\end{equation}
	where the factor $\lambda$ is a hyper-parameter used to balance the importance of components.
	In our experiments, $\lambda$ and $\alpha$ are set to $0.3$ and $0.6$ respectively according to the experiments on hyper-parameter analysis, as shown in ~\ref{sec:hp_analysis}.
 %
 

 \section{MSVR310 Benchmark}
	\label{sec:dataset}
	
	In this work, we release a new dataset called MSVR310 for multi-spectral vehicle Re-ID. 
	
	\subsection{Imaging Platform}
	
	In MSVR310, three different spectral modalities, RGB, NIR and TIR are captured for each sample.
	The RGB images are captured by two devices, a 360 D866 camera for day time and a Mi8 mobile phone camera for night time.
	All the NIR images are captured by the 360 D866 camera, which can be switched to the near infrared mode.
	The TIR image capture device is FLIR SC620 which contains a thermal infrared camera with the resolution of $640\times 480$.

	For each sample in our dataset, it is formed as a triplet constructed by three images from RGB, NIR and TIR respectively.
	We manually select bounding boxes for the targets in original captured images. 
	%
	
	
	\subsection{Data Setting and Statistics}
	
\begin{figure}[h]
		\centering
		\includegraphics[width=0.95\columnwidth]{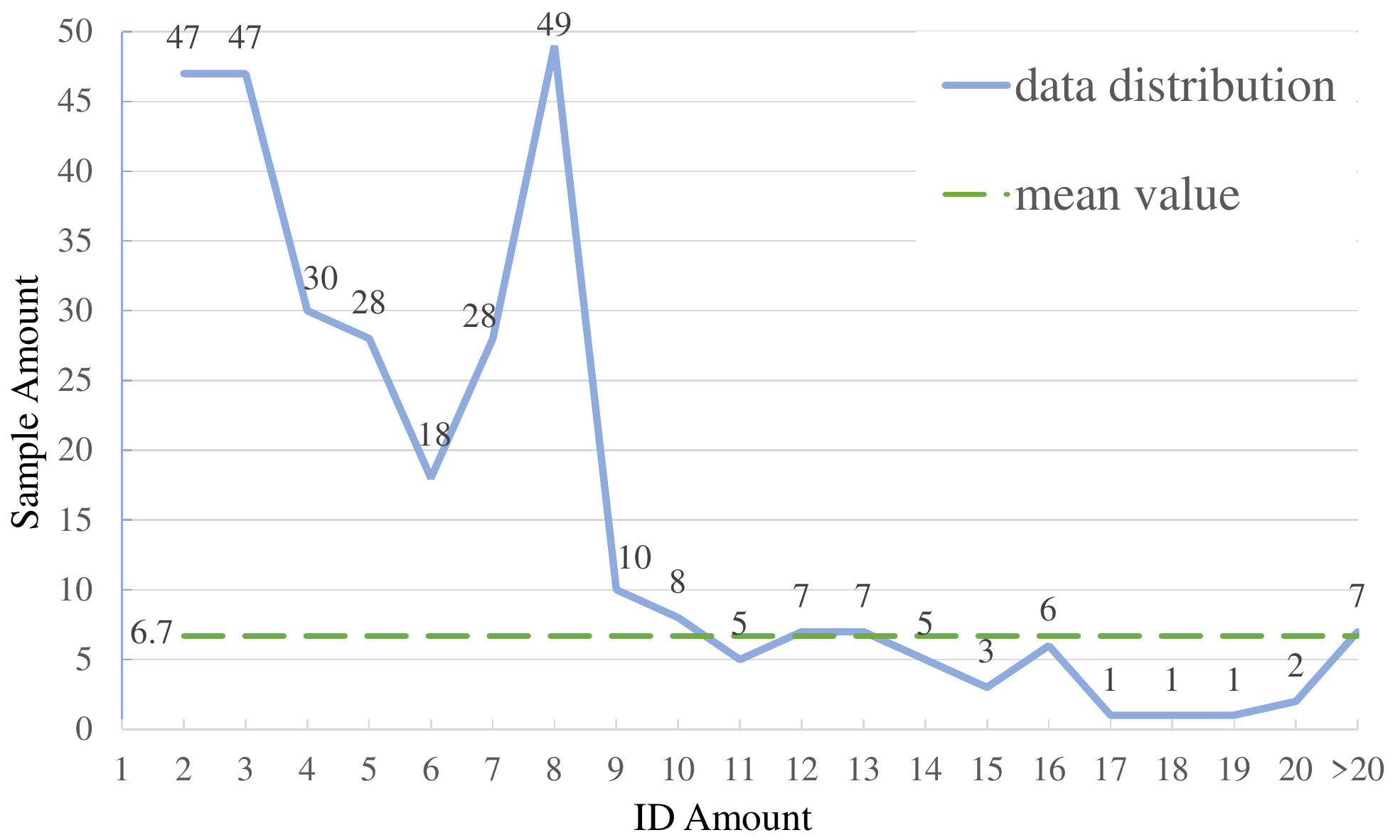}
		\caption{
			Distribution for number of identities across sample sizes. 
		}
		\label{fig:data_dstr}
	\end{figure}

	Our dataset contains 2087 samples from 310 vehicles and each sample is a triplet, which results in total 6261 images in our dataset.
	The number of image samples of each vehicle varies from 2 to 20. 
	We randomly select 155 vehicles with 1032 samples as the training set, while the rest 155 vehicles with 1055 samples as the gallery set. 
	We randomly select 52 vehicles with 591 samples from gallery set as query set.
	Each query identity has been captured at least twice with different time labels to support cross time matching.
	The data distribution is shown in Fig.~\ref{fig:data_dstr}.
	
\begin{figure}[t]
		\centering
		\includegraphics[width=0.95\columnwidth]{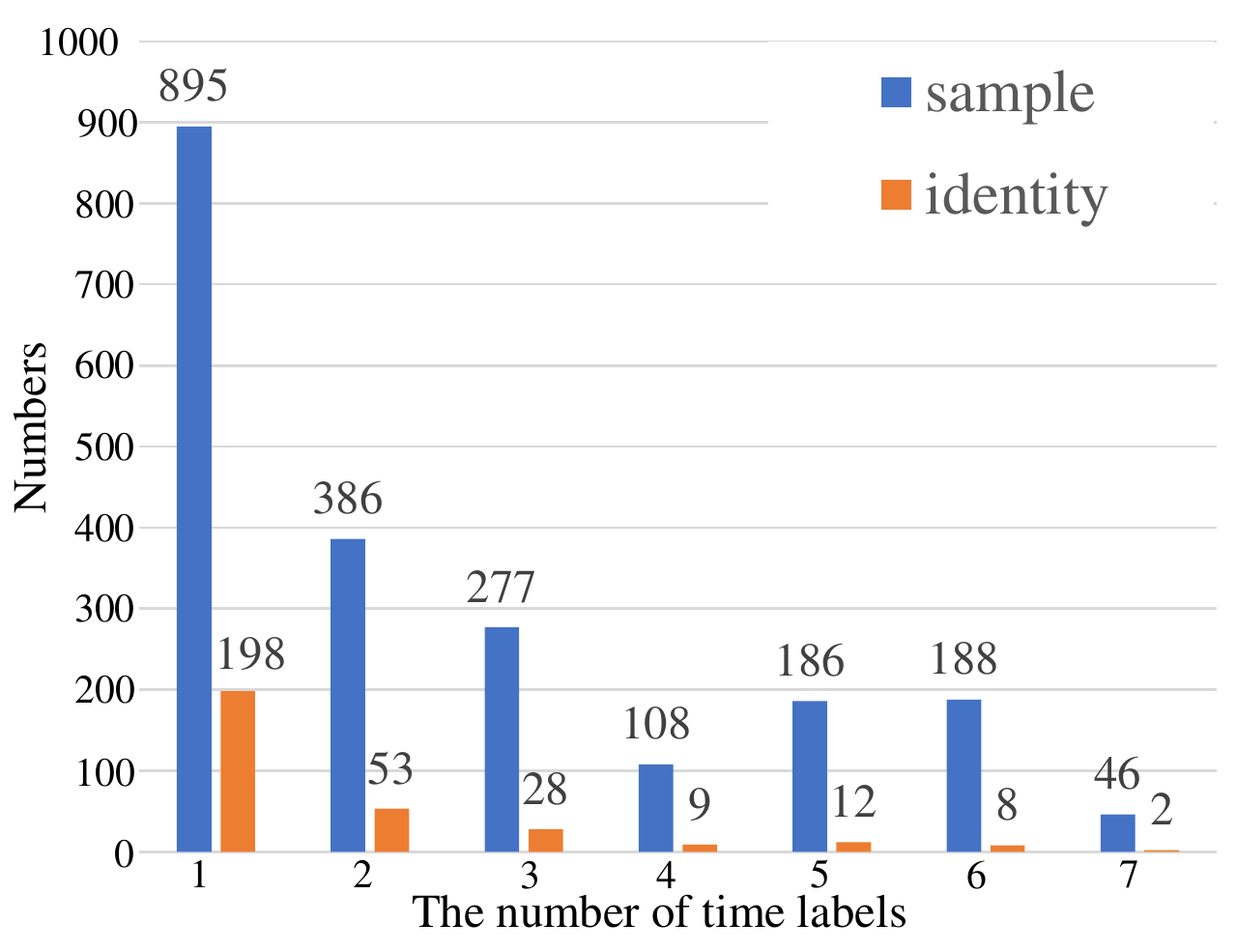}
		\caption{
			Distribution of samples and identities across the number of time labels in MSVR310.
		}
		\label{fig:scene_dstr}
	\end{figure}

\begin{figure*}[t]
		\centering
		\includegraphics[width=0.99\textwidth]{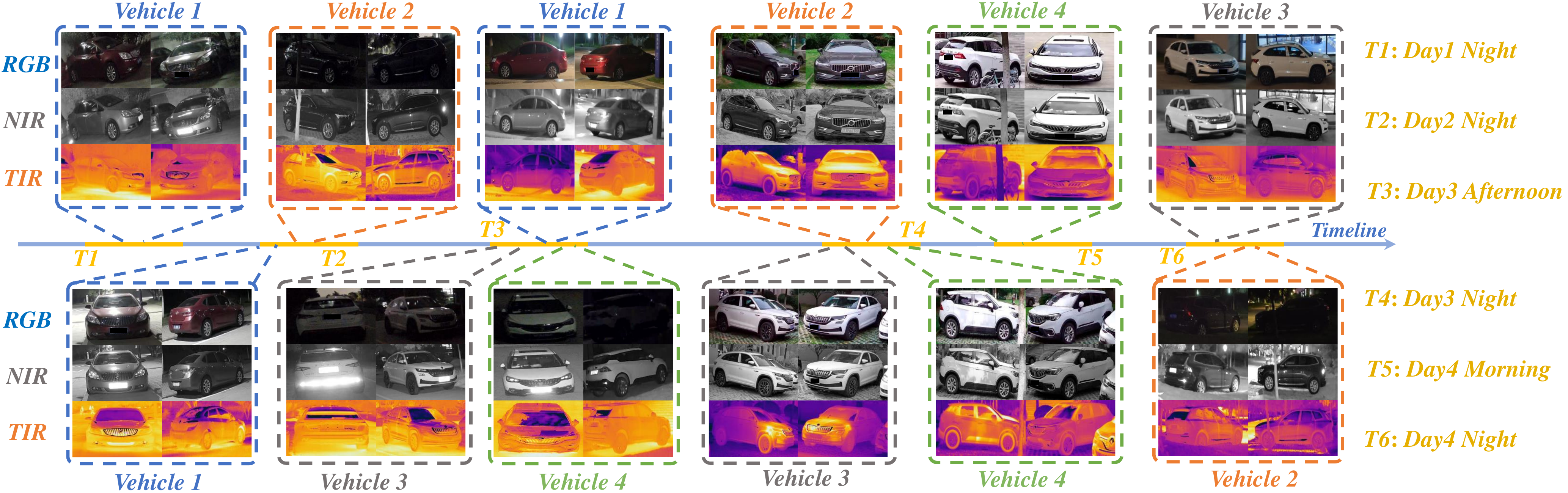}
		\caption{
			Illustration of four sample data in MSVR310. Images in box with the same color indicate the multi-modality samples of the same identity with different time labels.
		}
		\label{fig:data_show}
	\end{figure*}

	We annotate data with time labels according to their collection order along time.
	Fig.~\ref{fig:scene_dstr} demonstrates the distribution of the captured time.
	Fig.~\ref{fig:data_show} demonstrates some example images of four vehicles in MSVR310 along time labels.
	And each vehicle appears in various conditions with complex interference like strong illustration, reflection, shadow, color distortion and so on. 
	Thus, bad cases in a certain modality exist ubiquitously and intra-class appearance discrepancy is very significant in MSVR310. 
	The illumination disturbance in such degree is quite rare in existing works~\cite{Lou2019VERIWildAL,Liu2016LargescaleVR,Liu_2016_CVPR,li2020multi}.
	However, these disturbances represent differently in different modalities, and data across modalities are complementary in content against interference which requires for better utilization of multi-spectral data.
	
	\subsection{Difference from Previous Work}
	\label{sec:data diff}
	
	\begin{table}[]
	    \caption{
			Comparison of RGBN300, RGBNT100 and MSVR310, where '-' Denotes 'not Available'.
		}
  		\label{tab:dataset_comp}
  \scalebox{0.92}
  
		\begin{center}
			\scalebox{0.9}
            {
					\begin{tabular}{cc|c|c|c|c|c}
					\hline
					& Benchmark     & IDs    & Videos  & Modality & Views & Time Labels \\ \hline
					& RGBN300       & 300   & 4100 & R+N     & 8       & -           \\
					& RGBNT100      & 100   & 2070  & R+N+T   & 8       & -           \\
					& MSVR310       & 310   & 6261    & R+N+T   & 8       &  28         \\ \hline   
				    \end{tabular}	
			}
		\end{center}

	\end{table}

	Li \textit{et al.}~\cite{li2020multi} first propose two benchmarks multi-spectral vehicle Re-ID datasets RGBN300 and RGBNT100, as shown in Table~\ref{tab:dataset_comp}.
	First, although RGBN300 and RGBNT100 contain much more images than MSVR310, it is actually collected from 2070 short videos (690 videos for each modality) which leads to a bunch of similar frames.
	We construct MSVR310 in various environments such as large changes of illuminations, occlusions and weather by capturing high-quality images instead of videos.
	%
	%
	Second, MSVR310 is collected across long time spans which leads to rich collections of various environments and vehicles.
	These significantly increase the diversity and difficulty of our dataset.
	%
	%
	Third, although matching between samples in same identity and same viewpoint is not allowed in RGBN300 and RGBNT100~\cite{li2020multi}, environmental similarity among samples tends to raise easy matchings.
	Instead, MSVR310 introduces time labels to avoid easy matching.
	Matching between samples with the same identity and the same time label is forbidden in MSVR310, as shown in Fig.~\ref{fig:test_protocol}.
	This protocol effectively handles the easy matching problem and provides a more reliable evaluation.
	
	
	\section{Experiments}
	
	\subsection{Datasets and Evaluation Metrics}
	
	To evaluate the effectiveness of the proposed CCNet on our proposed multi-spectral vehicle Re-ID dataset and public dataset, we provide comprehensive experimental results in this section. 
	Due to there are only one public RGB-N-T image dataset RGBNT100 for the evaluation of multi-spectral vehicle Re-ID methods.
	We finally implement the experiments on MSVR310 and RGBNT100 following their own evaluation protocols.
	
	To ensure the fairness of experimental evaluation, we follow the commonly used Cumulative Matching Characteristic ($CMC$) curve and the mean Average Precision ($mAP$) for evaluation. 
	$CMC$ score reflects the retrieval precision, where $Rank-1$, $Rank-5$, $Rank-10$ scores are reported in our experiments. 
	$mAP$ measure the mean of all queries of average precision (the area under the Precision Recall curve), which reflects the recall and precision comprehensively.
	
	\subsection{Implementation Details}
	
	We use a strong baseline BoT~\cite{Luo2020ASB} which is modified from ResNet50~\cite{he2016deep} pretrained on ImageNet~\cite{deng2009imagenet} as our backbone and the implementation platform is Pytorch 1.0.1 with one NVIDIA GTX 1080Ti GPU.
	We use the Adam~\cite{kingma2014adam} optimizer to optimize our network with the initial learning rate as $3.5\times 10^{-4}$ which will be decayed to $3.5\times 10^{-5}$ and $3.5\times 10^{-6}$ at 300-th epoch and 550-th epoch respectively of total 1200 epochs.
	In training process, the input images are resized to $128\times 256$ and some data augmentation methods like random cropping, horizontal flipping and random erasing are used.
	We randomly select 8 identities which will provide 4 samples (12 images) by each one respectively as our training samples in each training mini-batch.
	In evaluation, we concatenate the features extracted after BNNeck~\cite{Luo2020ASB} from three parallel branches as final representation for a sample in the absence of additional instructions.
	
	\subsection{Evaluation on MSVR310 Dataset}
	\label{sec:eval_on_msvr310}
	
	\begin{table}[]
	    \caption{
			Experimental Comparison of the Effectiveness of Modalities between ResNet50 and CCNet on MSVR310 (in $\%$). In the Column Of Test Feature, \textit{R}, \textit{N} and \textit{T} Represents Features from Corresponding Spectrum (Branch) while '+' Denotes Feature Concatenating Operation. 
		}
		\begin{center}
        
        \scalebox{0.78}{
                \begin{tabular}{c|c|cccc}
				\hline
				\multicolumn{1}{l|}{Network} & \multicolumn{1}{l|}{Test Feature} & \multicolumn{1}{l}{mAP} & \multicolumn{1}{l}{Rank-1} & \multicolumn{1}{l}{Rank-5} & \multicolumn{1}{l}{Rank-10} \\ \hline
				\multirow{7}{*}{ResNet50}     
				& \textit{R}         & 20.0       & 29.9        & 49.9      & 61.6             \\
				& \textit{N}         & 17.8       & 28.9        & 51.3      & 62.8             \\
				& \textit{T}         & 11.9       & 23.2        & 37.4      & 46.4             \\
				& \textit{R + N}     & 23.6       & 36.7        & 57.0      & 66.2             \\
				& \textit{R + T}     & 22.6       & 35.4        & 54.7      & 63.5             \\
				& \textit{N + T}     & 21.4       & 37.2        & 56.3      & 64.3             \\
				& \textit{R + N + T} & 25.6       & 39.4        & 58.5      & 67.9             \\ \hline
				\multirow{7}{*}{CCNet}
				& \textit{R}         & 30.7       & 49.4        & 65.5      & 73.3             \\
				& \textit{N}         & 26.3       & 45.5        & 67.3      & 73.1             \\
				& \textit{T}         & 19.6       & 35.7        & 53.5      & 61.9             \\
				& \textit{R + N}     & 34.0       & 53.6        & 70.2      & 76.3             \\
				& \textit{R + T}     & 34.6       & 52.8        & 68.7      & 75.5             \\
				& \textit{N + T}     & 31.4       & 51.6        & 68.9      & 76.6             \\
				& \textit{R + N + T} & \textbf{36.4} & \textbf{55.2} & \textbf{72.4} & \textbf{79.7}      \\ \hline
			\end{tabular}}
		\end{center}
		\label{tab:res50vsCCNet}
	\end{table}

\begin{table*}[]
	\caption{
		Comparison to State-of-the-art Re-ID Methods on MSVR310 and RGBNT100 (in $\%$). The Best Three Scores Are Highlighted in \textcolor{red}{Red}, \textcolor{green}{Green}, and  \textcolor{blue}{Blue} respectively.
	}
	\begin{center}
		\scalebox{1}{
			\begin{tabular}{l|c|cccc|cccc}
				\hline
				\multirow{2}{*}{Models}
				& \multirow{2}{*}{Reference} 
				& \multicolumn{4}{c|}{MSVR310} 
				& \multicolumn{4}{c}{RGBNT100}
				\\ \cline{3-10} 
				& & mAP & Rank-1 & Rank-5 & Rank-10 
				& mAP & Rank-1 & Rank-5 & Rank-10
				\\ \hline
                DMML &  ICCV 2019            
				& 19.1 & 31.1 & 48.7 & 57.2 
				& 58.5 & 82.0 & 85.1 & 86.2 
				\\	
				Circle Loss & CVPR 2020
				& 22.7 & 34.2 & 52.1 & 57.2 
				& 59.4 & 81.7 & 83.7 & 85.2
				\\
				PCB & ECCV 2018
				& 23.2 & 42.9 & 58.0 & 64.6 
				& 57.2 & 83.5 & 85.6 & 87.9 
				\\
				MGN & ACM MM 2018
				& 26.2 & 44.3 & 59.0 & 66.8 
				& 58.1 & 83.1 & 85.6 & 88.0 
				\\
				Strong Baseline & CVPRW 2021
				& 23.5 & 38.4 & 56.8 & 64.8
				& \textcolor{red}{78.0} & \textcolor{blue}{95.1} & \textcolor{blue}{95.8} & \textcolor{blue}{96.4}
				\\
				HRCN & ICCV 2021
				& 23.4 & 44.2 & \textcolor{blue}{66.0} & \textcolor{green}{73.9}
				& 67.1 & 91.8 & 93.1 & 93.8 
				\\
				OSNet & ICCV2019
				& \textcolor{blue}{28.7} & \textcolor{blue}{44.8} & \textcolor{green}{66.2} & \textcolor{blue}{73.1}
				& 75.0 & \textcolor{green}{95.6} & \textcolor{blue}{97.0} & \textcolor{green}{97.4}
				\\ 
				AGW &  T-PAMI 2021
				& \textcolor{green}{28.9} & \textcolor{green}{46.9} & 64.3 & 72.3
				& 73.1 & 92.7 & 94.3 & 94.9
				\\
				TransReID & ICCV 2021
				& 26.9 & 43.5 & 62.4 & 70.7
				& \textcolor{blue}{75.6} & 92.9 & 93.9 & 94.6
				\\
				PFNet & AAAI 2021  
				& 23.5 & 37.4 & 57.0 & 67.3
				& 68.1 & 94.1 & 95.3 & 96.0 
				\\ 
				HAMNet & AAAI 2020
				& 27.1 & 42.3 & 61.6 & 69.5 
				& 74.5 & 93.3 & 94.3 & 95.2
				\\
                \textcolor{red}{PFD} & AAAI 2022
				&23.0   &39.9   &56.3   &64.0   
				& 67.5  & 92.6  &94.3   &96.5  
				\\
                \textcolor{red}{FED} & CVPR 2022
				&21.7   &37.4   &58.9   &67.3   
				&65.8   &91.7   &94.6   &96.3  
				\\
				
                \textcolor{red}{IEEE} & AAAI 2022
				&21.0   &41.0   &57.7   &65.0   
				&61.3 &87.8 &90.2 &92.1  
				\\
				\hline
				CCNet  &  OURS
				& \textcolor{red}{36.4} & \textcolor{red}{55.2} & \textcolor{red}{72.4} & \textcolor{red}{79.7}
				& \textcolor{green}{77.2} & \textcolor{red}{96.3} & \textcolor{red}{97.2} & \textcolor{red}{97.7}

				\\ \hline
			\end{tabular}
		}
	\end{center} 
	\label{tab:cmp_sota}
\end{table*}

We first evaluate our CCNet compared with the ResNet50 on MSVR310 dataset, as reported in Table~\ref{tab:res50vsCCNet}.
	%
	For fairness, we use the same implementation of ResNet50 from BoT~\cite{Luo2020ASB} for comparison, which is the same as the backbone of CCNet.
	Specifically, the results of ResNet50 are achieved by a multi-branch network constructed by three separated ResNet50 in which each branch handles data from a certain modality.
	The multi-modality branches are independent with no interaction with other branches, while CCNet simultaneously utilizes multi-spectral data in the training phase and thus achieves much better performance. 
	The \textit{R}, \textit{N} and \textit{T} in Table~\ref{tab:res50vsCCNet} represent the features used in test phase for distance computing from corresponding spectrum.
	Note that, we use all three modality data during the training phase.
	
	From Table~\ref{tab:res50vsCCNet} we can see, 
	i) First of all, none of the single spectrum achieves satisfactory performance due to the  complex lighting environments on MSVR310 dataset.
	In general, both RGB and NIR provide comparable reliable appearances thus lead to much better performance comparing to TIR.
	ii) Two spectrum scenarios significantly improve all the metrics than the single ones while the three spectrum scenarios further boost both performances of ResNet50 and CCNet.
	This strongly proves the effectiveness of the introduced multi-spectral data.
	iii) Our CCNet is superior to ResNet50 by a large margin while there are limited differences on network structure between CCNet and ResNet50.
	This strongly indicates the rightness of our discrepancy mitigating design and effectiveness of the proposed CdC loss and ALNU module.
	
	\subsection{\textcolor{red}{Evaluation on Different Backbones and Baselines}}
 {
    \begin{table}[]
	\caption{{\textcolor{red}{Plugin Our Key Components ALNU and CdC loss into Different Baselines and Backbones on MSVR310.}}}
	\begin{center}
		\scalebox{0.99}{
			\begin{tabular}{l|cccc}
				\hline
				\multirow{2}{*}{Methods}      & \multicolumn{4}{c}{MSVR310}      \\ \cline{2-5}
				& mAP  & Rank-1 & Rank-5 & Rank-10 \\ \hline
                SENet
				& 22.7 & 40.9 & 60.6 & 69.9
				\\
				+OURS
				& \textbf{29.5} & \textbf{47.0} & \textbf{67.7} & \textbf{73.8}
				\\ \hline

                InceptionV3
				& 23.1 & 43.7 & 59.7 & 68.5
				\\	
				+ OURS
				& \textbf{28.0} & \textbf{49.4} & \textbf{64.0} & \textbf{72.3} 
				\\ \hline

                {Desenet-121}
				& 35.8 & 54.1 & 71.2 & 81.2 
				\\	
				{+ OURS}
				& \textbf{38.7} & \textbf{58.5} & \textbf{76.5} & \textbf{82.2}
				\\ \hline

                {ResNet-101}
				& 25.2 & 38.9 & 58.5 & 68.2
				\\	
				{{+ OURS}}
				& \textbf{30.5} & \textbf{47.9} & \textbf{64.5} & \textbf{72.6}
				\\ \hline

                {ViT }
				& 30.8 & \textbf{49.9} & 66.2 & 72.1
				\\	
				{+$L_{CdC}$}
				& \textbf{34.4} & \textbf{49.9} & \textbf{69.4} & \textbf{78.7}
				\\ \hline
                
                MobileNetV2 & 22.5 & 37.6 & 53.6 & 64.1\\
				
				+ OURS
                
				& \textbf{24.0} & \textbf{43.5} & \textbf{59.2} & \textbf{70.1}\\
                
                \hline

                Strong Baseline              & 23.5 & 38.4   & 56.8   & 64.8    \\
				+ OURS & \textbf{25.6} & \textbf{47.0}   & \textbf{68.9}   & \textbf{74.6}    \\ \hline
				OSNet                         & 28.7 & 44.8   & 66.2   & 73.1    \\
				+OURS           & \textbf{30.3} & \textbf{50.3}   & \textbf{67.7}   & \textbf{75.3}    \\ \hline
				AGW                           & 28.9 & 46.9   & 64.3   & 72.3    \\
				+OURS            & \textbf{33.0} & \textbf{52.6}   & \textbf{69.5}   & \textbf{75.6}    \\ \hline
				TransReID                     & 26.9 & 43.5   & 62.4   & 70.7    \\
				+$L_{CdC}$              & \textbf{28.2} & \textbf{44.5}   & \textbf{62.3}   & \textbf{73.1}     \\
                \hline

			\end{tabular}	}
	\end{center}
	\label{tab:baseline_plugin}
\end{table}
    
    To validate the generality of our method, we integrate our CCNet into six backbones and four baselines including MobileNetV2~\cite{Sandler2018MobileNetV2IR}, SENet~\cite{Hu2020SqueezeandExcitationN}, InceptionV3~\cite{Szegedy2016RethinkingTI},
    { Desenet-121~\cite{huang2017densely}, ResNet-101~\cite{he2016deep}, ViT~\cite{dosovitskiy2020image}}, Strong Baseline~\cite{Huynh2021ASB}, OSNet~\cite{Zhou2019OmniScaleFL}, AGW~\cite{pami21reidsurvey} and TransReID~\cite{He2021TransReIDTO} as shown in Table~\ref{tab:baseline_plugin}.
    %
    Note that due to the conflict between convolution layer in ALNU and transformer structure, we only integrate the CdC loss into ViT~\cite{dosovitskiy2020image} and TransReID~\cite{He2021TransReIDTO}.
    %
	Generally speaking, after integrating our ALNU and CdC loss into the different baselines and backbones, all the metrics significantly improve, which indicates the generality of our method.
	%
	%
	}
	%
	%
	%

	\subsection{Comparison with State-of-the-art Methods}
	
	To validate the effectiveness of our method, we extend nine state-of-the-art single modality Re-ID methods including DMML~\cite{chen2019deep}, Circle loss~\cite{Sun2020CircleLA}, PCB~\cite{Sun2018BeyondPM}, MGN~\cite{Wang2018LearningDF}, Strong Baseline~\cite{Huynh2021ASB}, HRCN~\cite{Zhao_2021_ICCV},  OSNet~\cite{Zhou2019OmniScaleFL}, AGW~\cite{pami21reidsurvey} and TransReID~\cite{He2021TransReIDTO} to multi-modality version for comparison.
	At last, we compare our CCNet with the multi-spectrum vehicle Re-ID method HAMNet~\cite{li2020multi} and the multi-spectrum person Re-ID method PFNet~\cite{zheng2021robust}.
	Specifically, we train the single-modality methods on multiple spectral data respectively and then concatenate the final features from modalities of the same sample as the final representation. 
	The experimental comparison of these methods is shown in Table~\ref{tab:cmp_sota}.
	%

	First, all the methods perform much worse on MSVR310 than RGBNT100 which is caused by the huge challenge of the proposed MSVR310 dataset and our evaluation protocol which filters easy matchings caused by easy positive samples with same time label.
	The purposed CCNet beats all the comparison methods by a large margin on MSVR310, which strongly proves the effectiveness of our method.
	And on RGBNT100 which is much easier with fewer challenges and limited diversity, our method also achieves very competitive performance.
 \textcolor{red}{Transformer-based methods achieve superior performance in both person and vehicle Re-ID, due to their self-attention mechanism to simultaneously consider comprehensive local information for better global information learning. 
 However, in multi-modality Re-ID, the key challenge is to effectively explore the complementarity while suppressing the heterogeneity among different modalities. 
 Therefore, simply extending the transReID into the muli-modality task works overshadowed.}

	Second, as a first baseline multi-spectral vehicle Re-ID method, HAMNet~\cite{li2020multi} presents a simple network structure with considerable performance on three benchmark datasets, which proves its effectiveness on multi-spectral feature learning.
    {However, HAMNet mainly focuses on learning multi-modality feature relations and ignores the discrepancy in both sample and modality levels. While our CCNet mainly focuses on mitigating heavily intra-class and intra-modality discrepancies by introducing CdC loss and ALNU.}
	PFNet~\cite{zheng2021robust} is the first work for multi-spectral person Re-ID, while the local feature separation seems to be more suitable for person data than vehicle data. 

	\subsection{Ablation Study and Visualization}
	
	\begin{table}[]
	    \caption{Ablation Study on MSVR310 (in $\%$). \textcolor{red}{Note that $L_{CdC}$ = $L_{CdC_S}+L_{CdC_M}$}}.
		\begin{center}
			\begin{tabular}{l|cccc}
				\hline
				\multirow{2}{*}{Models} & \multicolumn{4}{c}{MSVR310}    \\ \cline{2-5} 
				& \multicolumn{1}{c}{mAP} & \multicolumn{1}{c}{Rank-1} & \multicolumn{1}{c}{Rank-5} & Rank-10 \\ \hline
				baseline                & 25.6 & 39.4 & 58.5 & 67.9 \\
				$+ALNU$                 & 29.4 & 47.2 & 66.0 & 74.3 \\
				$+L_{CdC_S}$            & 27.4 & 41.6 & 61.8 & 69.0 \\
				$+L_{CdC_M}$            & 31.4 & 48.6 & 65.1 & 73.6 \\
				$+ \textcolor{red}{L_{CdC_S}+L_{CdC_M}}$              & 33.7 & 51.8 & 68.2 & 76.0 \\
				$+L_{CdC}+ALNU$    & \textbf{36.4} & \textbf{55.2} & \textbf{72.4} & \textbf{79.7} \\ \hline
			\end{tabular}
		\end{center}
		\label{tab:ablation}
	\end{table}
	
	To verify the contributions of proposed components in our model, we implement the ablation study of several variants of CCNet on MSVR310 dataset, as reported in Table~\ref{tab:ablation}. 
	Note that \textcolor{red}{the sample center loss} $L_{CdC_S}$, \textcolor{red}{the modality center loss} $L_{CdC_M}$ and the adaptive layer normalization unit (ALNU) all make positive improvements on our baseline, which demonstrates the contributions of the corresponding modules.
	
	
	\begin{table}[]
	    \caption{
			Experimental Comparison with Different Normalizations and Losses on MSVR310 (in $\%$).
		}
		\begin{center}
			\begin{tabular}{c|cccc}
				\hline
				Methods & mAP & Rank-1 & Rank-5 & Rank-10 \\ \hline
				baseline& 25.6 & 39.4 & 58.5 & 67.9        \\ \hline

                + IN & 26.8 & 42.3 & 61.9 & 70.6  \\
				+ LN & 28.8 & 45.9 & \textbf{66.3} & 72.3  \\
				+ ALNU  & \textbf{29.4} & \textbf{47.2} & 66.0 & \textbf{74.3}       \\ \hline
				$+ L_{center}$ & 25.8 & 42.0 & 60.1 & 66.8 \\
				$+ L_{hc}$  & 30.5 & 48.7 & 64.8 & 72.1 \\
				$+ L_{CdC}$ & \textbf{33.7} & \textbf{51.8} & \textbf{68.2} & \textbf{76.0} \\ \hline
			\end{tabular}
		\end{center}
		\label{tab:norm_loss_cmp}
	\end{table}
	
	We verify the contribution of our ALNU module by comparing two conventional normalization operations, instance normalization (IN)~\cite{ulyanov2016instance} and layer normalization (LN)~\cite{ba2016layer} as shown in Table~\ref{tab:norm_loss_cmp}. 
	IN~\cite{ulyanov2016instance} is widely used in image style transfer by normalizing instance features in channel level directly. LN~\cite{ba2016layer} and ALNU both treat each feature as an entirety for normalization, however LN~\cite{ba2016layer} strictly enforces all features to follow the same mean value and variance while our ALNU dynamically learns the gain and bias factors which are more reasonable for complex data.
	We also verify the contribution of our CdC loss by comparing two widely used center-type losses, Center loss~\cite{2016A} and HC loss~\cite{2020Hetero} as shown in Table~\ref{tab:norm_loss_cmp}.
	Both HC loss and Center loss are implemented based on ResNet50 with same setting as our baseline. 
	We implement Center loss~\cite{2016A} to pull features within identity close regardless of modality.
	And HC loss~\cite{2020Hetero} is implemented to reduce the modality gap within identity.
	However, Center loss~\cite{2016A} is not good at handling the ubiquitous bad cases from a certain modality while HC loss~\cite{2020Hetero} ignores the discrepancy among intra-class samples in multi-modality situations.
	Both Center loss~\cite{2016A} and HC loss~\cite{2020Hetero} work overshadowed by our CdC loss which simultaneously constrains intra-class relations from both modality and sample aspects.
	This proves the validity and robustness of our CdC loss in multi-spectral vehicle Re-ID task.
	%
	
	\begin{figure}[t]
		\centering
		\includegraphics[width=1\columnwidth]{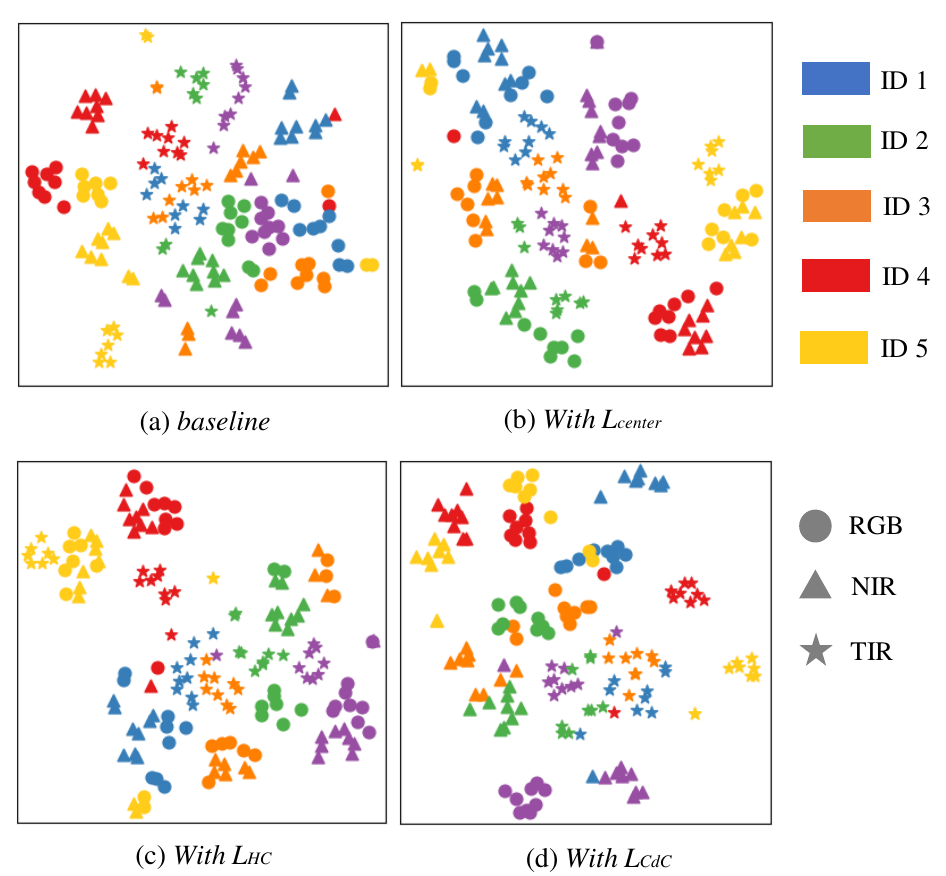}
		\caption{
			T-SNE~\cite{maaten2008visualizing} illustration of the feature distributions extracted by CCNet trained {(a) baseline, (b) \textcolor{red}{baseline} with $L_{Center}$, (c) \textcolor{red}{baseline} with $L_{HC}$ and (d) \textcolor{red}{baseline} with $L_{CdC}$.}
		}
		\label{fig:bcloss_compare}
	\end{figure}
	
    Fig.~\ref{fig:bcloss_compare} demonstrates the feature distribution comparison of the network trained with different losses.
	When training with the baseline as shown in Fig.~\ref{fig:bcloss_compare} (a), features from different modalities are mixed and hard to be separated by identity labels. 
	{After introducing the center loss $L_{Center}$, as shown in Fig.~\ref{fig:bcloss_compare} (b), features with same identity in a certain modality tend to be clustered together. However, the inter-modality gap of the same identity is still very large.}
	{As shown in Fig.~\ref{fig:bcloss_compare} (c), introducing the HC loss $L_{HC}$ can better eliminate the modality gap comparing with the center loss. However, some hard identities are still blended together such as the ID4, ID5 and ID6.}
	After introducing the proposed CdC loss $L_{CdC}$, as shown in Fig.~\ref{fig:bcloss_compare} (d), features from different modalities with same identity are constrained to follow stronger consistency in both sample and modality levels. 	
   

	\begin{figure}[t]
		\centering
		\includegraphics[width=0.99\columnwidth]{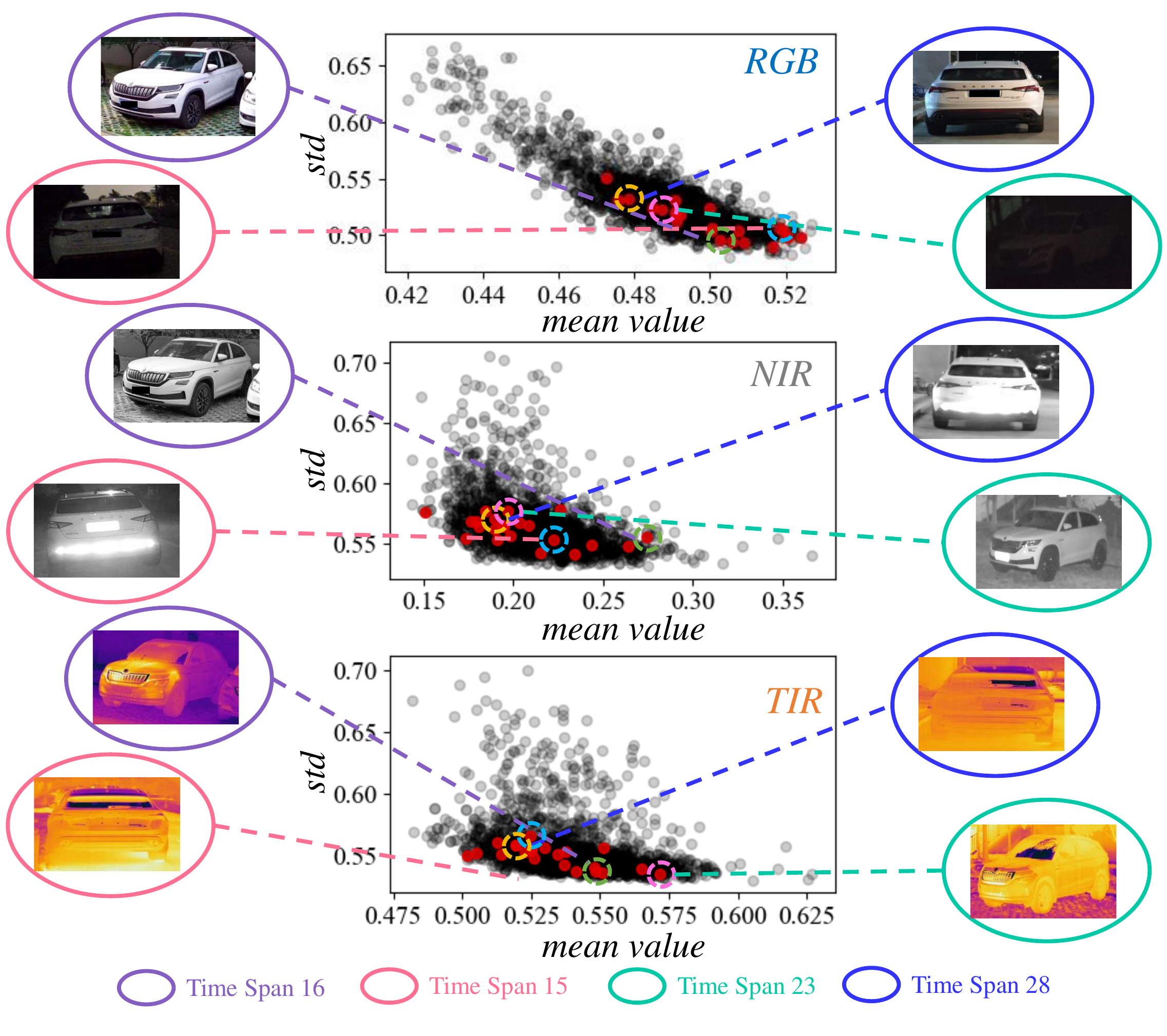}
		\caption{
			{Illustration of the feature distribution as shown in Fig.~\ref{fig:ALNU_cmp} after introducing ALNU. }
			Comparing to Fig.~\ref{fig:ALNU_cmp} we can obverse the distributional discrepancy is significantly mitigated.
		}
		\label{fig:ALNU_impact}
	\end{figure}
	
	Fig.~\ref{fig:ALNU_impact} demonstrates the distribution for multi-modality features after introducing ALNU. Compared with Fig.~\ref{fig:ALNU_cmp}, the ALNU pushes features to distribute with similar mean values and standard deviations to reduce the distributional variation.

\subsection{Comparison to Cross-modality Re-ID}

\begin{table*}[]
	\caption{{Comparison of State-of-the-art Cross-modality Re-ID Methods on  Reconstructed MSVR310.}}
	\begin{center}
		\scalebox{1}{
			\begin{tabular}{c|c|cccc|cccc}
				\hline
				\multicolumn{2}{c|}{\multirow{2}{*}{Method}} & \multicolumn{2}{c|}{RGB to TI}        & \multicolumn{2}{c|}{TI to RGB} & \multicolumn{2}{c|}{RGB to NI}        & \multicolumn{2}{c}{NI to RGB} \\ \cline{3-10} 
				\multicolumn{2}{c|}{}                        & mAP  & \multicolumn{1}{c|}{Rank-1} & mAP         & Rank-1        & mAP  & \multicolumn{1}{c|}{Rank-1} & mAP         & Rank-1       \\ \hline
                \multirow{3}{*}{Cross-Modality}     & LBA       & 10.4 & \multicolumn{1}{c|}{18.3}   & 11.2        & 19.0          & 22.5 & \multicolumn{1}{c|}{39.4}   & 21.7        & 39.6         \\
				& DDAG      & 11.9 & \multicolumn{1}{c|}{17.6}   & 13.5        & 21.3          & 23.0 & \multicolumn{1}{c|}{37.9}   & 22.5        & 39.1         \\
				& HCLoss   & 11.3 & \multicolumn{1}{c|}{20.0}   & 12.4        & 20.0          & 20.0 & \multicolumn{1}{c|}{37.9}   & 18.9        & 36.9         \\ 

                & \textcolor{red}{MPANet}   &12.6   &\multicolumn{1}{c|}{17.9}   &   11.52 &      15.7    &21.0 &\multicolumn{1}{c|}{35.5 }   &    20.1        &     37.9       \\
                & \textcolor{red}{MMN}   &6.48    &\multicolumn{1}{c|}{12.35}   & 5.2         &6.6          &20.4  &\multicolumn{1}{c|}{43.2}   &20.1         & 42.6          \\
                \hline
				Ours (Multi-Modality)            & CCNet     & \multicolumn{4}{c|}{mAP: {\bf 18.7} Rank-1: {\bf 29.1}}                        & \multicolumn{4}{c}{mAP: {\bf 29.4} Rank-1: {\bf 43.5}}                        \\ \hline
			\end{tabular}
		}
	\end{center}
	\label{tab:crossmodal}
\end{table*}

{
	To better evaluate the necessity of the multi-modality vehicle Re-ID, we compare our method with three state-of-the-art cross-modality Re-ID methods including LBA~\cite{park2021learning}, DDAG~\cite{ye2020dynamic}, HC Loss~\cite{2020Hetero}, MPANet~\cite{wu2021discover} and MMN~\cite{zhang2021MMN}.
	%
    Specifically, we reconstruct data in MSVR310 into cross-modality setting followed by the data splitting protocol in RegDB~\cite{Nguyen2017PersonRS} for the cross-modality evaluation.
	As shown in Table~\ref{tab:crossmodal},
	\textcolor{red}{due to the huge heterogeneity across modalities, all the five cross-modality Re-ID methods present inferior results}.
	CCNet achieves the distinct superior performance by utilizing both (RGB and TI/NI) modalities, which evidences that CCNet can simultaneously utilize the complementary information among the modalities and overcome the cross-modality heterogeneity.
}

{
	
}

\subsection{Evaluation on Random Modality Missing}
	
	\begin{figure}[t]
		\centering
		\includegraphics[width=0.99\columnwidth]{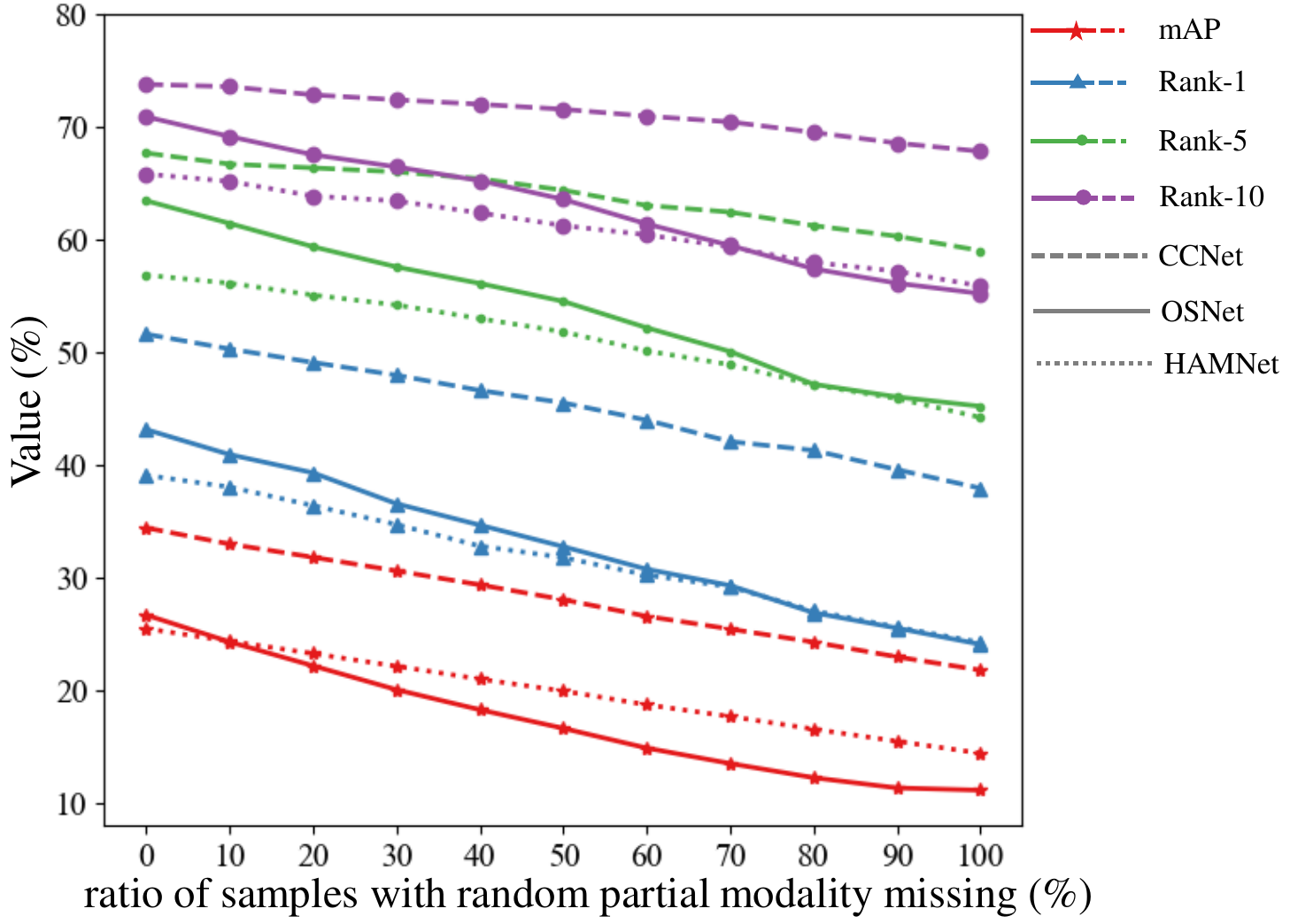}
		\caption{
			Performance changing of different methods in different ratio of samples with random partial modality missing on MSVR310.	{When a certain ratio of samples with random partial modality missing, the probability of missing one or two modalities is equal. }
		}
		\label{fig:rand_missing}
	\end{figure}
	
	To verify the generality of the proposed method and dataset in diverse real scenarios, we further evaluate CCNet in handling the missing modality issue.

	Specifically, we adjust the samples with a certain ratio of missing modalities in the test set for evaluation. The ratio indicates the probability of the samples with random partial (one/ two modalities in equal proportion) modality missing.
       {For example,
		ratio $r$\% indicates $r$\% of the samples suffer from random modality missing, which means half of them ($r/2$\%) randomly miss one modality while the rest half randomly missing two modalities. 
	}
	To overcome the sample feature misalignment caused by modality missing, we use geometric center of the existing modality/modalities as the final representation of the sample.
	
	%
	%
	
	In normal case without modality missing, CCNet extracts a final representation $f_{i,k}$ for sample $S_{i,k}$ (the $k^{th}$ sample for $i^{th}$ identity) where $f_{i,k}$ is a triplet of corresponding modality features. 
	To handle the modality missing case, we generate a binary triplet mask $T_{i,k}$ for $f_{i,k}$, to indicate whether the corresponding modality is missing or not.
	Then, the geometric center of sample can be formulated as:
	\begin{equation}
		\begin{aligned}
			{C^{\prime}_S}_{i,k} = \frac{1}{\sum T_{i,k}} \sum_{m=1}^{M} T^m_{i,k} f^m_{i,k},
		\end{aligned}
	\end{equation}
	where ${C^{\prime}_S}_{i,k}$ is the final representation of $f_{i,k}$.
	
	We evaluate the stability of our method in handling modality missing comparing with the representative multi-modality Re-ID method HAMNet~\cite{li2020multi} and the state-of-the-art single modality Re-ID method OSNet~\cite{Zhou2019OmniScaleFL}.
	All the experiments are evaluated based on the mean value of 10 random trials.
	%
	Fig.~\ref{fig:rand_missing} demonstrates the comparison performance against the ratio of samples with partial modality missing. 
	Generally speaking, CCNet consistently outperforms both HAMNet~\cite{li2020multi} and OSNet~\cite{Zhou2019OmniScaleFL} by a large margin.
	Even all the samples occur modality missing (when the ratio is 100\%), CCNet still achieves competitive performance which is comparable with the results at low missing ratio of HAMNet~\cite{li2020multi} and OSNet~\cite{Zhou2019OmniScaleFL}.
	%
	This verifies the stability of our method in handling the modality missing.
	Meanwhile, all the metrics drop as the missing ratio increases, especially for $mAP$ and $Rank-1$, which indicates the importance of complementary information of the multi-modality resources.
	As a state-of-the-art single modality Re-ID method, OSNet~\cite{Zhou2019OmniScaleFL} drops much faster than two multi-modality methods HAMNet~\cite{li2020multi} and CCNet, which indicates the advantage of fusing multi-modality information in the two multi-modality methods in handling modality missing issue.

	\subsection{Hyper-parameter Analysis}
	\label{sec:hp_analysis}
	
	\begin{table}[]
    	\caption{
			Hyper-parameter Analysis on MSVR310. (in $\%$)
		}
		\begin{center}
			\begin{tabular}{c|cccc}
				\hline
				Hyper-parameters & \multicolumn{4}{c}{MSVR310}                                                                                                      \\ \hline
				$\lambda$ ($\alpha = 0.6$)  & mA   & Rank-1 & Rank-5  & Rank-10  \\ \hline
				0.1              & 31.3 & 47.9 & 66.5 & 72.6 \\
				0.2              & 33.3 & 50.6 & 67.2 & 73.8   \\
				0.3              & \textbf{33.7} & \textbf{51.8} & \textbf{68.2} & \textbf{76.0}   \\
				0.4              & 33.4 & 50.9 & 67.0 & 74.6 \\
				0.5              & 33.0 & 50.6 & 67.7 & 74.3   \\
				0.6              & 32.8 & 50.4 & 66.8 & 73.3   \\
				0.7              & 32.7 & 50.1 & 66.2 & 73.8   \\ 
				0.8				 & 32.3 & 49.7 & 66.5 & 72.9 \\
				0.9    			 & 31.9 & 49.2 & 65.7 & 72.6 \\
				1.0              & 31.3 & 48.4 & 65.8 & 72.8 \\
				\hline
				$\alpha$ ($\lambda = 0.3$) & mAP   & Rank-1 & Rank-5  & Rank-10     \\ \hline
				0.1              & 32.6    & 48.6  & 66.2   & 74.0  \\
				0.2              & 33.5    & 51.6  & 67.2   & 76.0  \\
				0.3              & 33.1    & 50.4  & 66.9   & 75.5   \\
				0.4              & 33.4    & 50.1  & 67.1   & 76.4   \\
				0.5              & 33.1    & 50.1  & \textbf{68.9}    & \textbf{77.0} \\
				0.6              & 33.7    & \textbf{51.8}  & 68.2    & 76.0        \\
				0.7              & \textbf{33.8}   & 51.6   & 67.9    & 75.8     \\
				0.8              & 33.3    & 51.0  & 67.2   & 74.7   \\
				0.9              & 33.1    & 51.0  & 68.2   & 75.0     \\
				1.0              & 33.1    & 50.4  & 67.7   & 76.5     \\
				\hline
			\end{tabular}
		\end{center}
		\label{tab:param_analysis}
	\end{table}
	
	There are two hyper-parameters in our method, {\textit{e.g.}}, $\lambda$ in Eq.~\eqref{eq:L_total} which controls the importance of CdC loss in total loss and $\alpha$ in Eq.~\eqref{eq:L_cdc} which balances the strength of gradient along sample and modality directions in CdC loss.
	Large $\lambda$ may affect the inter-class discrimination ability provided by $L_{ce}$ and large $\alpha$ may break the balance between $L_{CdC-M}$ and $L_{CdC-S}$.
	Therefore, we vary $\lambda$ and $\alpha$ between 0.1 and 1.0 for the analysis.
	The analysis on diverse values of these two hyper-parameters is reported in Table~\ref{tab:param_analysis}.
	%
	It is clear that, our method achieves the top when $\lambda$ is set to $0.3$ while it is not sensitive to $\alpha$.
	We fix $\lambda$ and $\alpha$ as 0.3 and 0.6 for the best performance in our method.

	\section{Conclusion}
	
	
	In this work, we propose a novel end-to-end trained convolutional network named CCNet for robust multi-spectral vehicle Re-ID.
	CCNet contains a novel cross-directional center (CdC) loss to simultaneously overcome the problems of cross-modality discrepancy and intra-class individual discrepancy.
	Meanwhile, a simple yet effective module named adaptive layer normalization unit is designed to embed in CCNet to mitigate the distributional variation of intra-class features for robust feature learning.
	Furthermore, we create a high-quality benchmark dataset MSVR310 with diverse conditions and reasonable evaluation protocol.
	Comprehensive experiments on our benchmark dataset MSVR310 and the public dataset RGBNT100 validate the superior performance of our CCNet and the research value of the proposed benchmark dataset.







\section*{Data Availability Statement}
The data used to support the findings of this study are included in the paper.
\section*{Declaration of Competing Interest}
No potential conflict of interest was reported by the authors.

\section*{Author Statement}
Aihua Zheng: Conceptualization of this study and Methodology. 
Xianpeng Zhu: Investigation and Writing-
Original Draft. 
Zhiqi Ma: Validation and Visualization. Chenglong Li: Formal Analysis and Data Curation. 
Jin Tang: Resources and interpretation of data. 
Jixin Ma: Writing Review and Editing.

\section*{Acknowledgements}
This research is partly supported by the National Natural Science Foundation of China
(Grant No. 61976002) and the University Synergy Innovation Program of
Anhui Province (Grant Nos. GXXT-2022-036, GXXT-2020-051 and GXXT-
2019-025)

\bibliographystyle{cas-model2-names}
\bibliography{egbib.bib}


\bio{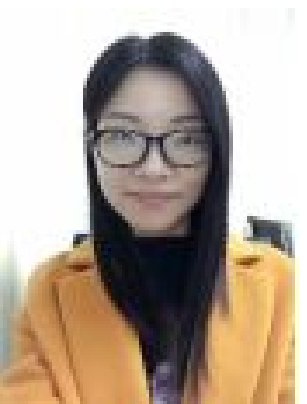}
{Aihua Zheng} received B.Eng. degrees and finished Master-Docter combined program in Computer Science and Technology from Anhui University of China in 2006 and 2008, respectively. And received Ph.D. degree in computer science from University of Greenwich of UK in 2012. She visited University of Stirling and Texas State University during June to September in 2013 and during September 2019 to August 2020 respectively. She is currently an Associate Professor and PhD supervisor at the School of Artificial Intelligence, Anhui University.
Her main research interests include vision based artificial intelligence and pattern recognition. Especially on person/vehicle re-identification, audio visual computing, and multi-modal intelligence.
\endbio

\bio{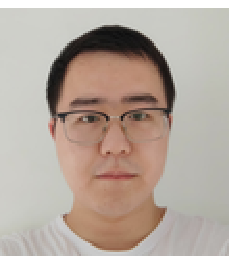}
{Xianpeng Zhu}
		received his B.Eng. degree in 2018 and is currently pursuing the M.Eng degree in the School of Computer Science and Technology, Anhui University, Hefei, China. His research interests include Computer Vision, Vehicle Re-identification, Multi-modal Intelligence and Deep Learning.
\endbio

\bio{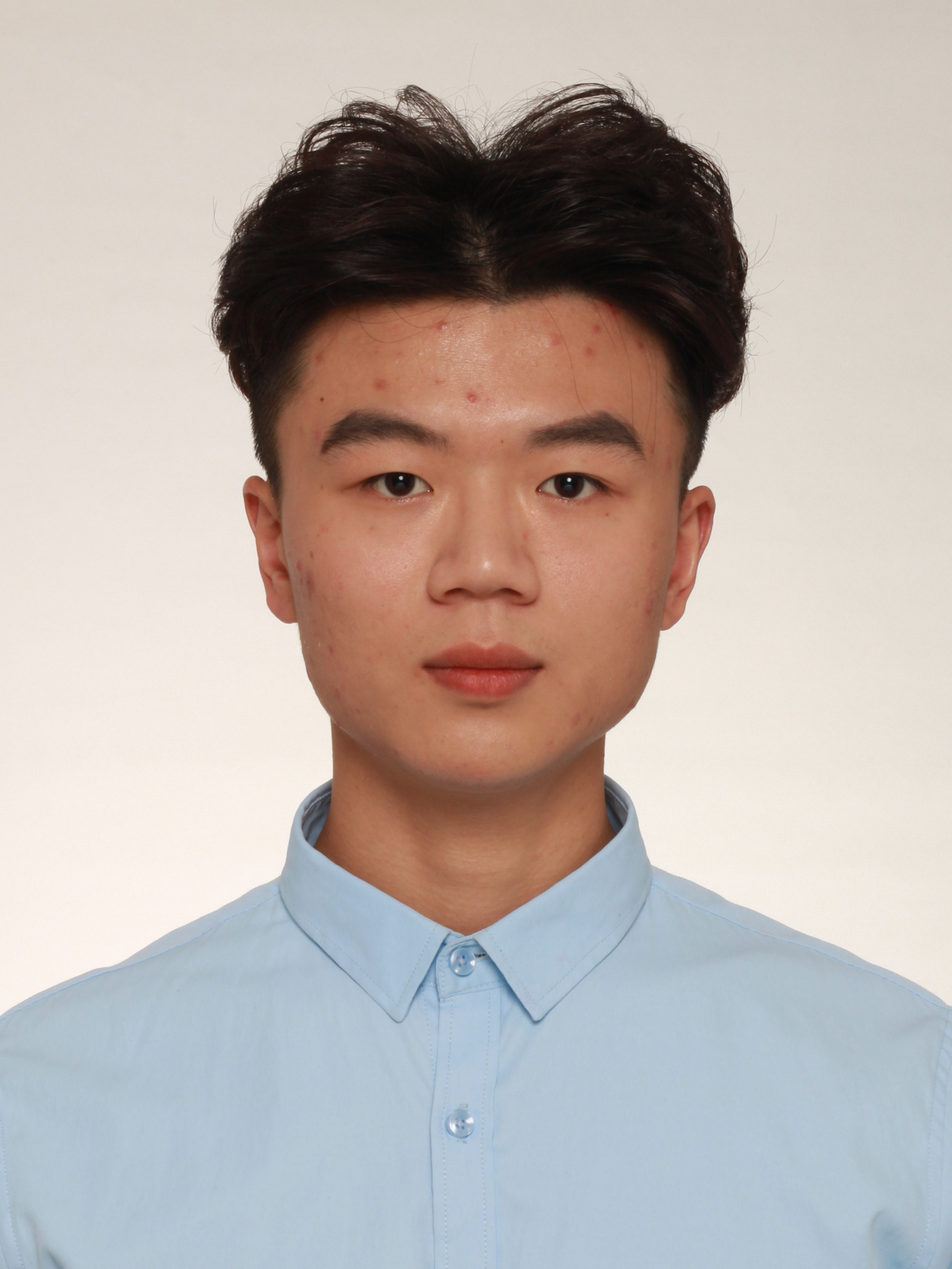}
{Zhiqi Ma}
		received his B.Eng. degree in 2021 and is currently pursuing the M.Eng degree in the School of Computer Science and Technology, Anhui University, Hefei, China. His research interests include Computer Vision, Multi-modal Intelligence and Vehicle Re-identification.
  \\
\endbio

\bio{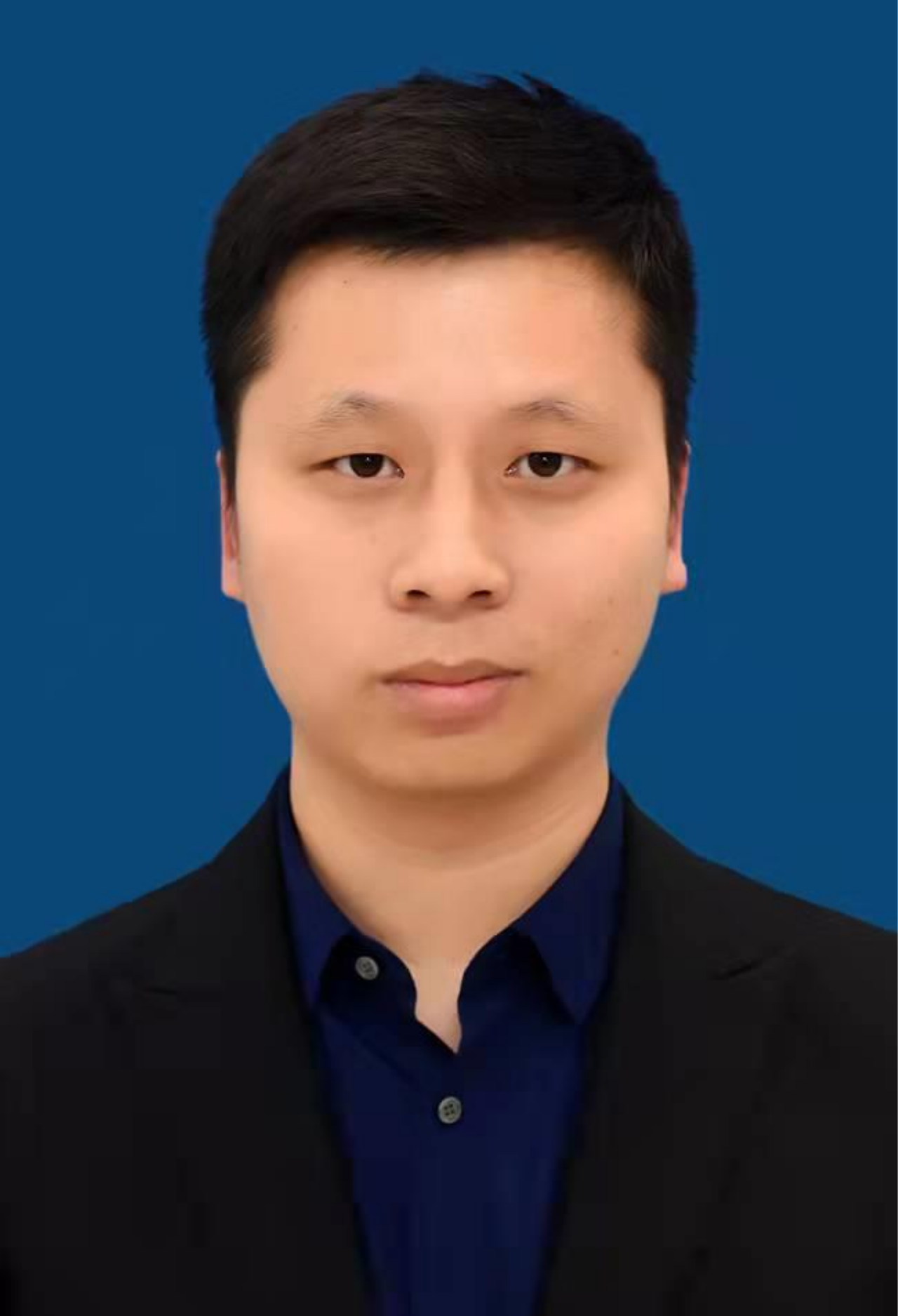}
{Chenglong Li}
		received the M.S. and Ph.D. degrees from the School of Computer Science and Technology, Anhui University, Hefei, China, in 2013 and 2016, respectively. From 2014 to 2015, he worked as a Visiting Student with the School of Data and Computer Science, Sun Yat-sen University, Guangzhou, China. He was a postdoctoral research fellow at the Center for Research on Intelligent Perception and Computing (CRIPAC), National Laboratory of Pattern Recognition (NLPR), Institute of Automation, Chinese Academy of Sciences (CASIA), China. 
		He is currently an Associate Professor and PhD supervisor at the School of Artificial Intelligence, Anhui University. His research interests include computer vision and deep learning. He was a recipient of the ACM Hefei Doctoral Dissertation Award in 2016.
\endbio

\bio{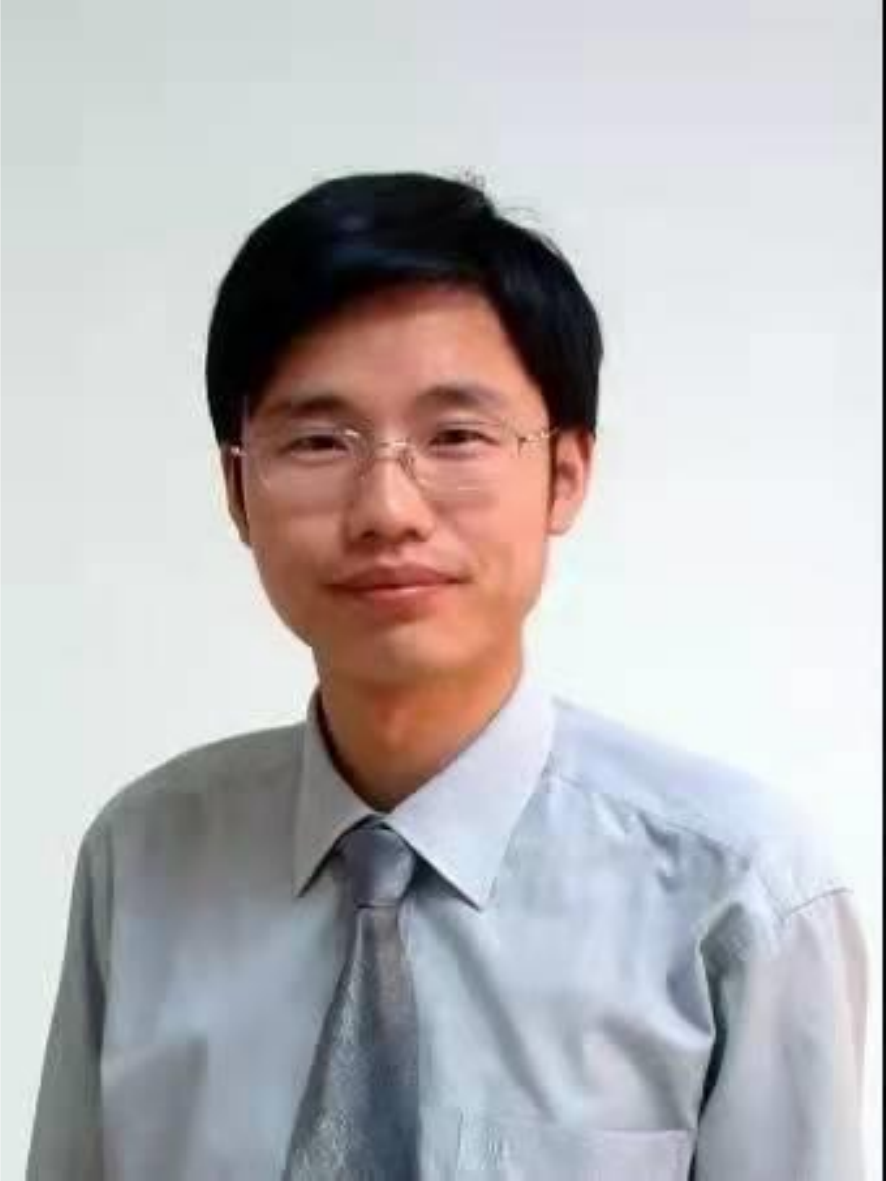}
{Jin Tang}
		received the B.Eng. degree in automation and the Ph.D. degree in Computer Science from Anhui University, Hefei, China, in 1999 and 2007, respectively. He is currently a Professor and PhD supervisor with the School of Computer Science and Technology, Anhui University. 
		His research interests include computer vision, pattern recognition, and machine learning.
\endbio

\bio{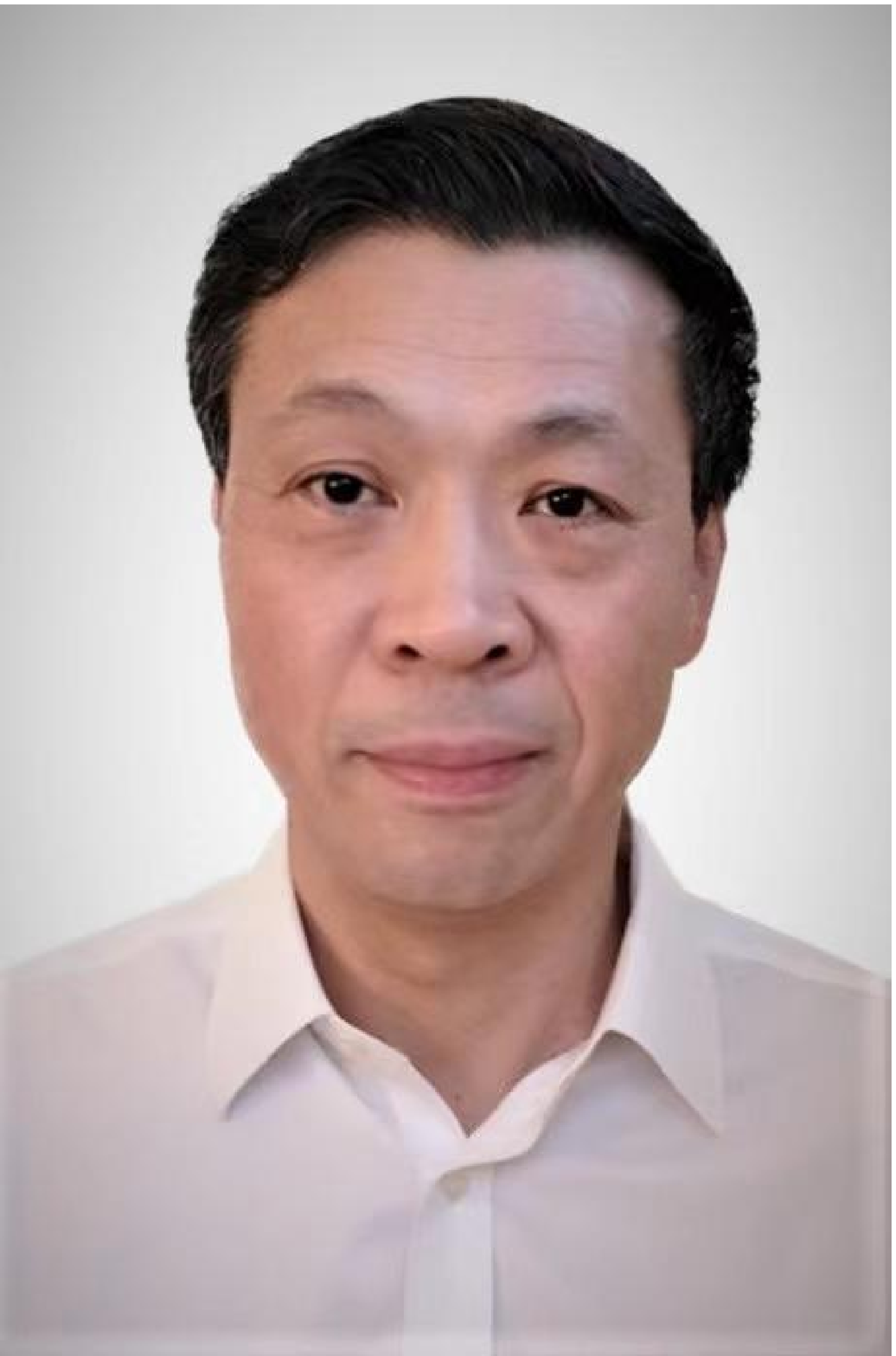}
{Jixin Ma}
	is a Full Professor and the Director of PhD/MPhil programme in the School of Computing and Mathematical Sciences, at University of Greenwich, U.K. Prof. Ma obtained his BSc and MSc of Mathematics in 1982 and 1988, respectively, and PhD of Computer Sciences in 1994. His research interests include Temporal Logic, Temporal Databases, Reasoning about Action and Change, Case-Based Reasoning, Pattern Recognition, Machine Learning and Information Security. 
\endbio

\end{document}